\documentclass[runningheads]{llncs}
\usepackage[T1]{fontenc}
\usepackage{graphicx}
\usepackage{amsmath, amsfonts, amssymb}
\usepackage{caption, subcaption}
\usepackage{wrapfig, lipsum}
\usepackage{cleveref}
\usepackage{booktabs}
\usepackage{multirow}
\usepackage{enumitem}

\begin{document}
\title{An Empirical Study of the Influence of Adversarial Fine-Tuning on Compressed Neural Networks}
%
%
\author{Hallgrimur Thorsteinsson\inst{1} \and
Valdemar J Henriksen\inst{1} \and
Daniel I R Cruz\inst{1} \and Raghavendra Selvan\inst{1}\orcidID{0000-0003-4302-0207} \and Tong Chen\thanks{corresponding author}\inst{1}\orcidID{0000-0003-3752-9971} }
\authorrunning{H. Thorsteinsson et al.}
%
\institute{Department of Computer Science, University of Copenhagen, Denmark \email{\{hlq592,hcl907,vfd797\}@alumni.ku.dk}, \email{\{raghav,toch\}@di.ku.dk}}
\maketitle              
\begin{abstract}
    As deep learning (DL) models are increasingly being integrated into our everyday lives, ensuring their safety by making them robust against adversarial attacks has become increasingly critical. DL models have been found to be susceptible to adversarial attacks by introducing small, targeted perturbations to disrupt the input data. Adversarial training has been presented as a mitigation strategy that can result in more robust models. This adversarial robustness comes with additional computational costs required to design adversarial attacks during training. The two objectives -- adversarial robustness and computational efficiency -- then appear to be in conflict with each other. In this work, we explore the effects of neural network compression on adversarial robustness. We specifically explore the effects of fine-tuning on compressed models, and present the trade-off between standard fine-tuning and adversarial fine-tuning. Our results show that {\em adversarial fine-tuning} of compressed models can yield large improvements to their robustness performance. We present experiments on several benchmark datasets showing that adversarial fine-tuning of compressed models can achieve robustness performance comparable to adversarially trained models, while also improving computational efficiency. \footnote{Source code is available here: \url{https://github.com/saintslab/Adver-Fine}.}

\end{abstract}
\section{Introduction} \label{sec:intro}
The growing computational cost of large-scale deep learning (DL) models is concerning because of their increasing energy use and the resulting carbon emissions~\cite{sevilla2022compute,strubell2019energy}. Many solutions are being explored to improve efficiency at different stages of a DL model’s life cycle~\cite{bartoldson2023compute}. Compressing neural networks to reduce computational demands during training and deployment has been very successful. Extreme compression methods, such as neural network pruning with high levels of weight sparsity~\cite{hoefler2021sparsity,lecun1989optimal} and quantization using low-precision weights or activation maps, often cause little to no performance loss~\cite{dettmers2022eight,hubara2016binarized}. The usual trade-off in model compression is between efficiency and test performance. However, it is still unclear how compressing neural networks affects other properties, such as adversarial robustness, which is safety-critical applications such as autonomous driving and medical diagnosis.~\cite{biggio2013evasion,huang2017safety}.

Adversarial training~\cite{Aleksander2018adv,Alexey2016adv,tramer2018ensemble} is a standard approach to improving the robustness of DL models. It works by adding carefully designed noise to the training data (known as adversarial examples or attacks) and then training the model on this modified data~\cite{Explaining2015Goodfellow}. Creating these adversarial examples adds computational costs compared to standard training~\cite{Shafahi2019Adversarial,Wong2020Fast}. This added cost conflicts with the goal of improving efficiency, for example through model compression. The central question this work addresses is: {\em Can adversarial robustness be \underline{efficiently} achieved for compressed neural networks?}

We investigate the possibility of simultaneously achieving computational efficiency and adversarial robustness. We show that adversarial fine-tuning of already compressed models can reach performance comparable to uncompressed models that are adversarially trained from scratch, resulting in compounded efficiency gains. To this end, we make the following contributions.
\begin{enumerate}
    \item Present adversarial fine-tuning of compressed models as a means of achieving robustness efficiently;
    \item Conduct a comprehensive empirical study using model pruning and quantization on multiple benchmark datasets, with and without adversarial fine-tuning.
\end{enumerate}

\section{Related Works}

\textbf{Model compression:} Model compression in machine learning (ML) refers to reducing the size of an ML model while maintaining its performance as much as possible. Smaller models require fewer computational resources and generally have lower inference times, making them more efficient for deployment in resource-constrained environments.

Model pruning~\cite{gorodkin1993quantitative} is a technique that removes model parameters with little influence on test performance. Pruning can be categorized as unstructured or structured. In unstructured pruning, individual parameters are removed, while in structured pruning, groups of parameters (such as weights of a kernel or transformer layers) are removed in one operation. 

Quantization~\cite{hubara2018quantized,wangexploring} reduces the precision of model weights or intermediate activation maps~\cite{eliassen2024activation} from high to lower precision (for example, from 32-bit to fewer bits). Quantization can yield large reductions in memory usage and inference time, and it can be adapted to specific hardware devices for acceleration.

In addition to pruning and quantization, knowledge distillation has shown potential for compressing large networks into smaller ones~\cite{hinton2015distilling}. For large-scale models, the gains from distillation are substantial, as observed in vision~\cite{chen2017learning} and large language models~\cite{sanh2019distilbert}.

Another approach to compressing neural networks is tensor factorization of model weights~\cite{novikov2015tensorizing}. Techniques such as tensor trains~\cite{oseledets2011tensor} have been used to factorize network weights, resulting in a considerable reduction in the overall number of trainable parameters~\cite{yin2021towards}. Using knowledge distillation in conjunction with tensor decomposition has been shown to be even more effective, since it allows the factorized tensor cores to relearn representations that may be lost during factorization~\cite{wangexploring}. 

\noindent\textbf{Effects of model compression:} The primary goal of model compression is to improve efficiency by reducing the number of parameters or the memory consumption while preserving downstream test performance. Recent work has shown drastic reductions in parameter counts~\cite{wangexploring} or extreme quantization~\cite{dettmers2022eight} while retaining performance comparable to uncompressed models. There are no formal theories that fully explain how extreme compression is possible, though recent attempts, such as the lottery ticket hypothesis, speculate about the existence of sub-networks within larger networks that can be recovered through compression~\cite{frankle2018the}. 

Beyond the trade-off between test performance and efficiency, compression may also affect other model properties. For example, even if overall performance is comparable to the original model, certain subsets of data may suffer disproportionately high errors, leading to unexpected effects on fairness~\cite{hooker2020characterising,ramesh2023comparative,stoychev2022effect}. On the other hand, recent work shows that knowledge distillation can improve fairness~\cite{chai2022fairness,jung2021fair} and adversarial robustness~\cite{maroto2022benefits} in DL models.

\noindent\textbf{Robustness-aware model compression:} To mitigate the negative effects of compression on adversarial robustness~\cite{jordao2021effect}, several approaches incorporate robustness as an additional regularization term, such as Lipschitz regularization~\cite{lin2018defensive}, during compression, attempting to optimize for both efficiency and robustness simultaneously~\cite{Goldblum2020Adversarially,Gui2019Model,ye2019adversarial}. Robustness-aware pruning techniques~\cite{jian2022pruning,vikash2020HYDRA} have also been proposed and are useful in safety-critical and resource-constrained applications. It has also been shown that adversarial fine-tuning~\cite{jeddi2020simple} of a standardly trained model can significantly improve adversarial robustness without the need for full adversarial training. In this work, rather than jointly optimizing for efficiency and robustness as in~\cite{lin2018defensive}, we propose a simpler yet effective approach: first compress, then adversarially fine-tune. Our experimental results show consistent behavior across various datasets and models, as demonstrated in \Cref{tab:final_quant}.

\section{Methods for Model Compression and Adversarial Robustness} \label{sec:methods}

The standard process of model compression usually consists of three steps: (1) train a large over-parameterized model, which may overfit to some extent; (2) apply compression techniques to reduce the size of the trained model while preserving its performance as much as possible; (3) fine-tune the compressed model to recover some of the lost performance and ensure it performs well on the target task. In this work, we focus on two compression methods: structured pruning and quantization.

\subsection{Structured pruning}
We consider $\ell_1$-norm based filter pruning \cite{li2017pruning}, which is a simple but effective way of structured pruning for convolutional neural networks (CNNs), {which is a widely used pruning methods}. Suppose we have an input of shape $c_{\text{in}} \times h_{\text{in}} \times w_{\text{in}}$ where $c_{\text{in}}$ is the number of input channels, and $h_{\text{in}} \times w_{\text{in}}$ is the height and width of the input features. A convolutional layer, denoted by $\mathbf{F}_j$, is a mapping that takes an input of shape $c_{\text{in}} \times h_{\text{in}} \times w_{\text{in}}$ to an output of shape $c_{\text{out}} \times h_{\text{out}} \times w_{\text{out}}$, which is realized by $c_{\text{out}}$ many filters of shape $c_{\text{in}} \times k \times k$:
$$\mathbf{F} = [\mathbf{F}_1, \ldots, \mathbf{F}_{c_{\text{out}}}]: \mathbb{R}^{c_{\text{in}} \times h_{\text{in}} \times w_{\text{in}}} \longrightarrow \mathbb{R}^{c_{\text{out}} \times h_{\text{out}} \times w_{\text{out}}}.$$
Each filter consists of $c_{\text{in}}$ kernels of shape $k \times k$ that maps individually the corresponding channel in the input of shape $h_{\text{in}} \times w_{\text{in}}$ to an output of shape $h_{\text{out}} \times w_{\text{out}}$, depending on padding and stride parameters:
$$\mathbf{F}_j = \sum_{i = 1}^{c_{\text{in}}} \mathbf{F}_{i,j}: \mathbb{R}^{c_{\text{in}} \times h_{\text{in}} \times w_{\text{in}}} \longrightarrow \mathbb{R}^{h_{\text{out}} \times w_{\text{out}}},$$
where each filter $\mathbf{F}_{i,j}$ acts on the $i$-th channel of the input.

Now compute the $\ell_1$-norm of each filter $\mathbf{F}_j$, and denote by $s_j = \|\mathbf{F}_j\|_1 = \sum_{i = 1}^{c_{\text{in}}} \|\mathbf{F}_{i,j}\|_1$. Depending on the sparsity ratio of pruning, we sort the filters by the values $s_j$ and leave out the bottom filters with the lowest $\ell_1$-norm. Note that each time a filter is removed, the output features of the next layer and the corresponding kernels in the next layer are removed. In this way, the new filters are obtained for both the current layer and the next layer. We do the pruning process for both standardly and adversarially trained models, which is also called post-train pruning. Note that structured pruning can also be applied to other model architectures, such as transformers, by replacing filters with corresponding model weights.

\subsection{Quantization}
A quantization scheme consists of a quantizer that maps a real number, $r$, to an integer: $q(r) = \lfloor r / s \rceil - z$, and the dequantizer: $\hat{r} = s (q(r) + z)$, where $s \in \mathbb{R}$ is called a \emph{scaling factor}, and $z \in \mathbb{Z}$ is called a \emph{zero point}. This procedure is also called uniform quantization, since the quantized values are uniformly distributed due the rounding operator $\lfloor \cdot \rceil$. 

The scaling factor $s$ is usually of form $s = (\beta - \alpha) / (2^b - 1)$, where $[\alpha, \beta]$ is the clipping range and $b$ is the bit width of quantization, a.k.a. $b$-bit quantization. A common choice of $\alpha$ and $\beta$ is the min-max value of the real number $r$, i.e., $\alpha = \min (r)$ and $\beta = \max (r)$. In this case, $-\alpha$ is not necessarily equal to $\beta$, hence we call it asymmetric quantization. We can also set $-\alpha = \beta = \max (|\min(r)|, |\max(r)|)$, which is called symmetric quantization. Both of them have their advantages: asymmetric quantization usually gives tighter clipping range, and symmetric quantization simplifies the computations. However, using symmetric quantization wastes half of the precision on ReLU activation, because none of the negative values in the quantization grid is used. For these reasons we use symmetric quantization for weight and asymmetric quantization for activation maps in this work. 

We use \emph{Post-Training Quantization (PTQ)} \cite{nagel2021white} throughout this work. PTQ is a method which can be easily applied and it is efficient compared to, e.g., \emph{Quantization Aware Training (QAT)} \cite{jacob2018quantization}. As the name suggests, PTQ takes a pre-trained model and quantizes it. The method may be data-free, but can also be applied with a small unlabeled dataset to adjust the quantization. The implementation that we use, takes care of the adjustment of calibrating scaling factors and zero points. This ensures that the resulting quantization ranges strike a favorable balance between rounding and scaling errors.

\subsection{Adversarial Training {and Fine-tuning}}
Consider the $n$-dimensional Euclidean space $\mathbb{R}^n$ endowed with norm $\|\cdot\|$. For $p \ge 1$ and $\mathbf{x} \in \mathbb{R}^n$, the $\ell_p$-norm is defined as $\|\mathbf{x}\|_p = (\sum_{i = 1}^n |x_i|^p)^{1/p}$ if $p < \infty$, and $\|\mathbf{x}\|_p = \max_i |x_i|$ if $p = \infty$. Given a finite dataset $\mathcal{S} = \{(\mathbf{x}_i, y_i)\}_{i = 1}^N \subseteq \mathbb{R}^{n+1}$, where each data $(\mathbf{x}_i, y_i)$ is assumed to be i.i.d. sampled from some unknown distribution $\mathcal{D}$, we are trying to learn a function $f: \mathbb{R}^n \rightarrow \mathbb{R}$ that maps all $\mathbf{x}_i$ to $y_i$. 

Assume the functions are taken from some hypothesis space $\mathcal{H}$, we define the generalization loss of $f \in \mathcal{H}$ as $L (f) = \mathbb{E}_{(\mathbf{x}, y) \sim \mathcal{D}} [l(f(\mathbf{x}), y)]$, where $l: \mathbb{R}^{2} \rightarrow \mathbb{R}_+$ is a loss function. The empirical loss of $f$ is defined as
\begin{equation} \label{eq:standard}
    \hat{L}_{\mathcal{S}} (f) = \frac{1}{N} \sum_{i = 1}^N l(f(\mathbf{x}_i), y_i).    
\end{equation}
A \emph{standard model} is a function in $\mathcal{H}$ that minimizes the empirical loss, i.e., $f_{st} \in \arg\min_{f \in \mathcal{H}} \hat{L}_{\mathcal{S}} (f)$. 

For perturbation {radius} $\varepsilon > 0$ and norm $\|\cdot\|$, the adversarial loss of $f$ is defined as
$L (f, \varepsilon) = \mathbb{E}_{(\mathbf{x}, y) \sim \mathcal{D}} [\max_{\|\delta\| \le \varepsilon} l(f(\mathbf{x} + \delta), y)],$
and the empirical adversarial loss is defined as
\begin{align} \label{eq:adv}
    \hat{L}_{\mathcal{S}} (f, \varepsilon) = \frac{1}{N} \sum_{i = 1}^N \max_{\|\delta\| \le \varepsilon} l(f(\mathbf{x}_i + \delta), y_i).
\end{align}
A \emph{robust model} is a function in $\mathcal{H}$ that minimizes the empirical adversarial loss, i.e., $f_{rb} \in \arg\min_{f \in \mathcal{H}} \hat{L}_{\mathcal{S}} (f, \varepsilon)$. {In the following, we use the term \emph{adversarial training} to refer to minimizing the adversarial loss from a randomly initialized model until convergence, and \emph{adversarial fine-tuning} to refer to minimizing the adversarial loss from a pre-trained model for only a few epochs.}

For a model $f \in \mathcal{H}$, the \emph{test performance} of $f$ over dataset $\mathcal{S}$ is given by the test accuracy on clean data: $\# \{(\mathbf{x}_i, y_i): f(\mathbf{x}_i) = y_i\} / N$, and the \emph{robustness performance} of $f$ over $\mathcal{S}$ is computed by the test accuracy on all possible adversarial perturbations: $\# \{(\mathbf{x}_i, y_i): f(\mathbf{x}_i + \delta) = y_i, \; \forall \; \|\delta\| \le \varepsilon\} / N$. However, solving the maximization problem in \cref{eq:adv} is usually difficult, therefore evaluating the exact robustness performance of a model is not tractable. In practice, we use a simple and common strategy, called {\em Projected Gradient Descent} (PGD) \cite{madry2018towards}, to obtain a lower bound of the maximum. In fact, with PGD, the gradient descent is performed over the negative loss function: at step $t$, we update $\mathbf{x}^t$ by 
$$\mathbf{x}^{t+1} = \text{Proj}_{\mathbf{B} (\mathbf{x}_i, \varepsilon)} (\mathbf{x}^t + \alpha \cdot \text{sign} (\nabla_{\mathbf{x}} l(f(\mathbf{x}), y)) |_{\mathbf{x} = \mathbf{x}^t}),$$
where $\mathbf{B} (\mathbf{x}_i, \varepsilon)$ is the ball around $\mathbf{x}_i$ with radius $\varepsilon$ and some norm $\|\cdot\|$, $\alpha$ is the step size of the PGD iteration, and $\text{Proj}_{\mathbf{B} (\mathbf{x}_i, \varepsilon)}$ is the projection map. 

Denote by $\delta_i^{pgd}$ the adversarial perturbation obtained by PGD, then each $\mathbf{x}_i + \delta_i^{pgd}$ serves as an adversarial attack. The robustness performance of $f$ is estimated (and in fact, upper bounded) based on the number of correct predictions on the worst-case perturbation, i.e., $\# \{(\mathbf{x}_i, y_i): f(\mathbf{x}_i + \delta_i^{pgd}) = y_i\} / N$. For conciseness, we follow the notations in Table~\ref{tab:notation} throughout the paper.

\begin{table}[ht]
    \vskip -0.3in
    \centering
    \scriptsize
    \renewcommand{\arraystretch}{1.3}
    \caption{Summary of notations for models with different training, compression, and fine-tuning methods.} \label{tab:notation}
    \makebox[1 \textwidth][c]{
        \begin{tabular}{ll}
            \toprule
            \textbf{Notation} & \textbf{Description} \\
            \midrule
            $f_{st}$ (resp. $f_{rb}$)  & standard (resp. robust) model \\
            $f^{c}$  & any compressed model \\
            $f^{p}$ (resp. $f^{q}$) & pruned (resp. quantized) model \\
            $f^{p}_{st}$ (resp. $f^{p}_{rb}$) & pruned standard (resp. robust) model \\
            $f^{q}_{st}$ (resp. $f^{q}_{rb}$) & quantized standard (resp. robust) model \\
            $\mathcal{T}_{st} (f)$ (resp. $\mathcal{T}_{ad} (f)$) & standardly (resp. adversarially) fine-tuned model \\
            $\mathcal{T}_{st} (f^p_{st})$ (resp. $\mathcal{T}_{st} (f^q_{st})$) & pruned (resp. quantized) standard model with standard fine-tuning \\
            $\mathcal{T}_{st} (f^p_{rb})$ (resp. $\mathcal{T}_{st} (f^q_{rb})$) & pruned (resp. quantized) robust model with standard fine-tuning \\
            $\mathcal{T}_{ad} (f^p_{st})$ (resp. $\mathcal{T}_{ad} (f^q_{st})$) & pruned (resp. quantized) standard model with adversarial fine-tuning \\
            $\mathcal{T}_{ad} (f^p_{rb})$ (resp. $\mathcal{T}_{ad} (f^q_{rb})$) & pruned (resp. quantized) robust model with adversarial fine-tuning \\
            \bottomrule
        \end{tabular}
    }
    \vskip -0.3in
\end{table}

\section{Data \& Experiments}
{\bf Data and models:} All main experiments were performed on multiple datasets of varying complexity: MNIST, FashionMNIST, SVHN, CIFAR-10, CIFAR-100, and TinyImageNet, using WideResNet (WRN) and Vision Transformer (ViT). In addition, an 8-layer CNN with 6 convolutional blocks and ResNet-18~\cite{he2016deep} were used for a subset of hyperparameter tuning experiments.


The experiments were performed on a single NVIDIA Titan RTX with 16GB of GPU memory using PyTorch~\cite{paszke2019pytorch}. We used the Neural Network Intelligence (NNI) library~\cite{nni2021} for quantization and pruning, following the setup of \cite{Tutorial} for the two adversarial attacks: PGD and AutoAttack. For quantized models, we used the training framework from \cite{jacob2018quantization}, which employs integer-only arithmetic during inference and floating-point arithmetic during training.


 


\noindent{\bf Experiments:} We use two sets of experiments to build the empirical evidence in this work. We conduct hyperparameter studies and configure the main experiments using {\em small-scale experiments} that rely on the simple 8-layered CNN trained on FashionMNIST and ResNet-18 trained on CIFAR10. Guided by these results, we expand our study to the {\em large-scale experiments} comprising Wide ResNet and Vision transformer models, that are trained on MNIST, FashionMNIST, SVHN, CIFAR-10, CIFAR-100, and TinyImageNet datasets. 
At a high level, the main experiments can be categorized as:
\begin{enumerate}[leftmargin=*]
    \item Full model training: $f_{st}, f_{rb}$;
    \item Standard fine-tuning of compressed models: $\mathcal{T}_{st}(f_{st}^c), \mathcal{T}_{st}(f_{rb}^c)$;
    \item Adversarial fine-tuning of compressed models: $\mathcal{T}_{ad}(f_{st}^c), \mathcal{T}_{ad}(f_{rb}^c)$.
\end{enumerate}
The superscript $c$ denotes either pruning or quantization. See Table~\ref{tab:notation} for an overview of the notations used. The results and trends from these experiments are discussed in detail in the next section.

\noindent{\bf Hyperparameters:} Both standard and adversarial training were run for a fixed 20 epochs using stochastic gradient descent (SGD) without momentum. The learning rate was set to $10^{-1}$ for the first four epochs and then reduced to $10^{-2}$. All PGD attacks were run for 20 iterations with an attack learning rate of $10^{-2}$. We set the adversarial perturbation $\varepsilon$ for the $\ell_{\infty}$ norm to $8/255$ for all datasets. For the large-scale experiments, we evaluate all compressed, standard, and robust networks using AutoAttack~\cite{croce2020reliable} with APGD-CE and APGD-DLR. The fine-tuning hyperparameters after compression vary across experiments (see~\Cref{app:hyperparm} for details).

\section{{Results}} \label{sec:results}   

\subsection{Small-scale Experiments}

\noindent{\bf Pruning ratio and quantization precision:} To choose the appropriate compression level for structured pruning and quantization, we conducted a comprehensive sweep of compression values using the small-scale CNN model on FashionMNIST. We used $[0.1, 0.2, \dots, 0.9]$ for sparsity ratios and $[\texttt{INT16}$, $\texttt{INT8}$, $\texttt{INT4}$, $\texttt{INT2}$, $\texttt{INT1}]$ for quantization precision, following the general settings in the literature~\cite{kuzmin2024pruning}.

\begin{figure}[t]
    \centering
    \begin{subfigure}{0.49\textwidth}
        \centering
        \includegraphics[width=\linewidth]{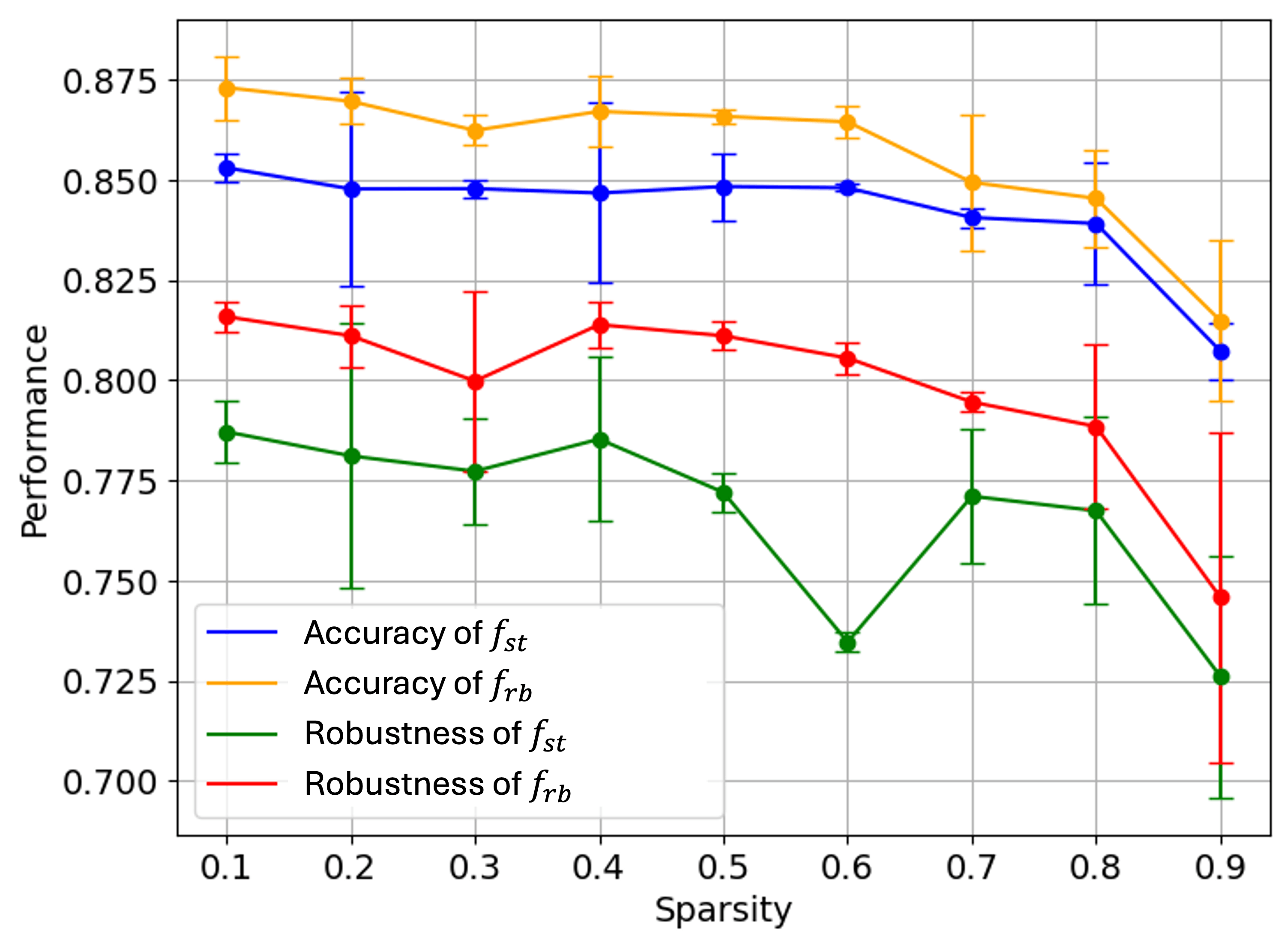}
        \caption{$\ell_1$-norm pruning on F-MNIST}
        \label{fig:suba_Clevels}
    \end{subfigure}
        \centering
    \begin{subfigure}{0.49\textwidth}
        \centering
        \includegraphics[width=\linewidth]{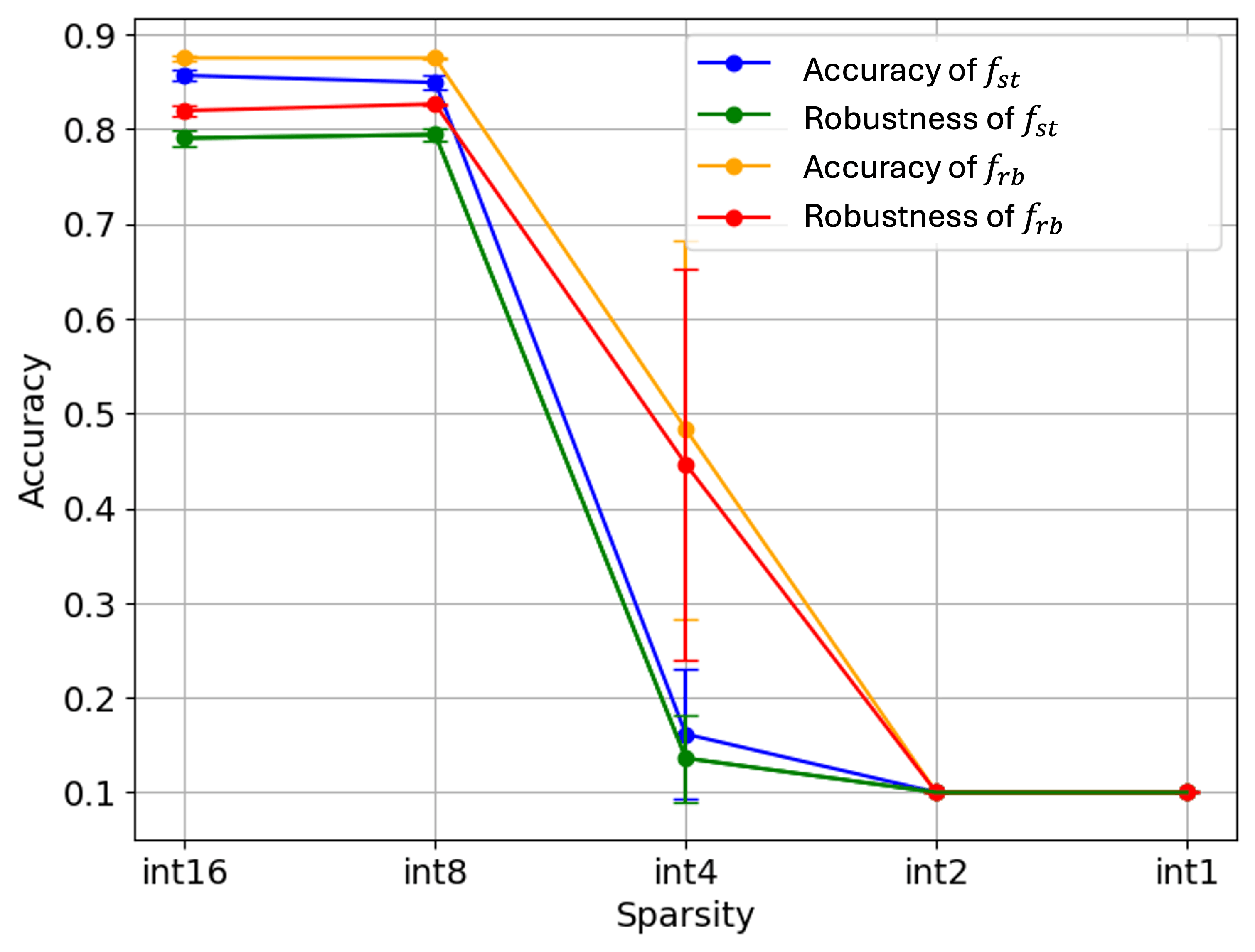}
        \caption{PTQ on F-MNIST}
        \label{fig:subc_Clevels}
    \end{subfigure}
    \caption{{Performance of compressed models on FashionMNIST with adversarial fine-tuning $\mathcal{T}_{ad} (\cdot)$. We perform $\ell_1$-norm pruning (\Cref{fig:suba_Clevels}) and post-train quantization (\Cref{fig:subc_Clevels}) on standard and robust models. In each subfigure, the horizontal axis shows the level of compression performed on the model, and the vertical axis shows the performance. Each model was trained three times and averaged out, error bars show the standard deviation between runs. Note that the scaling of performance are different for pruning and quantization.}}
    \label{fig:compressionlevels}
    \vskip -0.15in
\end{figure}

Using these configurations, the models were trained under four settings: standard training without fine-tuning, adversarial training without fine-tuning, standard fine-tuning $\mathcal{T}_{st}(\cdot)$, and adversarial fine-tuning $\mathcal{T}_{ad}(\cdot)$. We then selected the levels where the test performance of compressed standard models with adversarial fine-tuning, $\mathcal{T}_{ad}(f^c_{st})$, was comparable. This led to our choice of an $80\%$ sparsity ratio versus \texttt{INT8} quantization, as shown in \Cref{fig:compressionlevels}. We argue that this is a fairer way to set compression levels between methods such as pruning and quantization, rather than relying on arbitrary choices that could give undue advantage to one method, as in~\cite{li2017pruning}. \footnote{It is important to distinguish between reducing numerical precision and inducing sparsity, since they affect a model in fundamentally different ways. For example, converting weights from INT16 to INT8 halves the number of bits used to store each weight, but all weights are still present and used in computation. In contrast, 50\% weight sparsity removes or zeroes out half of the weights. Therefore, halving precision does not necessarily correspond to inducing 50\% sparsity, even if the memory reduction may appear similar.}

\begin{table}[h]
    \scriptsize
    \centering
    \caption{{Performance of the compressed models on Fashion-MNIST (using simple CNN) and CIFAR10 (using ResNet-18) datasets without fine-tuning (``None'' column), with standard fine-tuning $\mathcal{T}_{st} (\cdot)$ and $\mathcal{T}_{ad} (\cdot)$ for adversarial fine-tuning. We consider $\ell_1$-pruning and \texttt{INT8} post-train quantization.}}
    \label{tab:fintuning}
    \makebox[1 \textwidth][c]{
        \begin{tabular}{ccccc}
            \toprule
            \multirow{3}{*}{\textbf{Dataset}} & \multirow{3}{*}{\textbf{Method}}  & \multicolumn{3}{c}{\textbf{Fine-Tuning}} \\
            \cmidrule{3-5}
            && \multicolumn{1}{c}{None} & \multicolumn{1}{c}{$\mathcal{T}_{st}(\cdot)$} & \multicolumn{1}{c}{$\mathcal{T}_{ad}(\cdot)$}\\
            \midrule
            \multirow{5}{*}{{FashionMNIST}}&{$f_{st}^p$} &  33.21$\pm$3.96/00.33$\pm$0.64 & \textbf{88.94$\pm$1.03}/00.28$\pm$0.64 & 83.91$\pm$1.53/\textbf{76.74$\pm$2.33}\\
            \cmidrule{2-5}
            \multirow{5}{*}{{(Simple CNN)}}&{$f_{st}^q$} &  \textbf{90.40$\pm$0.16}/11.72$\pm$2.71 & 90.07$\pm$0.56/7.00$\pm$1.44 & 84.93$\pm$0.71/84.93$\pm$0.71\\
            \cmidrule{2-5}
            &{$f_{rb}^p$} &  16.68$\pm$4.57/14.40$\pm$5.47 & \textbf{87.71$\pm$0.40}/16.26$\pm$5.45 & 84.53$\pm$1.21/\textbf{78.84$\pm$2.04}\\
            \cmidrule{2-5}
            &{$f_{rb}^q$} & 87.90$\pm$0.34 /\textbf{82.80$\pm$0.14} & \textbf{89.54$\pm$0.47}/25.66$\pm$10.61 & 87.51$\pm$0.04/ 82.65$\pm$0.11\\
            \midrule
            \multirow{5}{*}{{CIFAR10}}&{$f_{st}^p$} &  86.68$\pm$0.01/00.00$\pm$0.00 & \textbf{89.74$\pm$0.33}/00.00$\pm$0.00 & 81.98$\pm$0.71/\textbf{56.99$\pm$0.11}\\
            \cmidrule{2-5}
            \multirow{5}{*}{{(ResNet-18)}}&{$f_{st}^q$} &  88.23$\pm$0.00/0.09$\pm$0.00 & \textbf{90.75$\pm$0.16}/0.01$\pm$0.00 &84.21$\pm$0.93/\textbf{60.03$\pm$0.67} \\
            \cmidrule{2-5}
            &{$f_{rb}^p$} &  74.95$\pm$0.67/35.31$\pm$0.77 & \textbf{89.31$\pm$1.33}/03.26$\pm$0.32 &83.18$\pm$0.88/\textbf{57.13$\pm$0.42}\\
            \cmidrule{2-5}
            &{$f_{rb}^q$} & 85.64$\pm$0.32 /\textbf{58.08$\pm$0.92} & \textbf{90.75$\pm$0.72}/03.98$\pm$0.59 &84.31$\pm$0.22/57.23$\pm$0.63\\
            \bottomrule
        \end{tabular}
    }
    \vskip -0.15in
\end{table}

\noindent{\bf Role of fine-tuning:} To isolate the effects of fine-tuning after compression, we first report the performance of compressed models without and with fine-tuning, as shown in~\Cref{tab:fintuning}. This analysis is conducted using the small CNN model on FashionMNIST and ResNet-18 on CIFAR-10. We observe that both test performance and robust accuracy of compressed models drop significantly without fine-tuning (``Fine-Tuning: None'' column). 

We also examine the influence of standard fine-tuning on robust models, $\mathcal{T}_{st}(f_{rb}^p), \mathcal{T}_{st}(f_{rb}^q)$ (column under $\mathcal{T}_{st}(\cdot)$ in Table~\ref{tab:fintuning}). For compressed robust models, standard fine-tuning helps recover test performance but leads to a significant reduction in robustness. Furthermore, we conduct a similar analysis using adversarial fine-tuning. Across the board, we observe an improvement to the robust accuracy (last column in Table~\ref{tab:fintuning}).

These small-scale experiments motivate highlight that adversarial fine-tuning helps. We expand this observation to the large-scale experiments in the next section.

\subsection{Main Experiments}

{\bf Full model training:} In this setting, no model compression is performed, and it serves as our baseline to assess the impact of compression (and adversarial fine-tuning) on robustness. We perform standard and adversarial training by minimizing \cref{eq:standard} and \cref{eq:adv}, respectively.\footnote{Accuracy values are reported as ``$*/*$'', with the left value corresponding to standard accuracy and the right to robust accuracy under AutoAttack.} Full models (without compression) are used for all datasets, and the results are reported in \Cref{tab:final_quant}, first column (under the label $f_{st}$). We clearly observe that the standard models, $f_{st}$, have poor adversarial robustness across all datasets. Adversarial training produces robust models and improves robustness (under the label $f_{rb}$) across the board, with a small drop in test performance, which is expected when performing adversarial training.

\noindent{\bf Standard fine-tuning of compressed models:} As discussed in \Cref{sec:methods}, we use (1) PTQ with symmetric quantization for weights combined with asymmetric quantization of activation maps, and (2) structured pruning of model weights as our preferred compression techniques. Furthermore, we apply standard fine-tuning to the compressed models for a fixed number of epochs. This allows us to compare these compression methods in a manner similar to~\cite{kuzmin2024pruning}, but with a focus on adversarial robustness. We first examine the influence of standard fine-tuning, $\mathcal{T}_{st}(\cdot)$, on standard models $f_{st}^p$ and $f_{st}^q$. These results are reported under the columns $\mathcal{T}_{st}(f_{st}^q)$ and $\mathcal{T}_{st}(f_{st}^p)$ in Table~\ref{tab:final_quant}, respectively.


For quantized models with standard fine-tuning $\mathcal{T}_{st}(f_{st}^q)$, it is noteworthy that the test performance of the standard models does not improve further. This can be explained by the fact that PTQ in its standard formulation typically does not involve fine-tuning. 

In contrast, standard fine-tuning of the pruned model, $\mathcal{T}_{st}(f_{st}^p)$, recovers test performance comparable to that of the full models $f_{st}$. However, standard fine-tuning does not recover any adversarial robustness for either pruned or quantized standard models, which is consistent with expectations.



\noindent{\bf Adversarial fine-tuning of compressed models:} One of the main questions considered in this work is how to improve both adversarial robustness and computational efficiency of DL models. To study this, we now turn to adversarially fine-tuning the compressed models. These results are reported under $\mathcal{T}_{ad}(\cdot)$ in~\Cref{tab:final_quant} for both quantization and pruning, with the baseline (uncompressed) model robust accuracy in the column $f_{rb}$.

The first thing to note is that the robust accuracy after adversarial training varies between datasets. This is shown in column $f_{rb}$ in Table~\ref{tab:final_quant}. All the models were adversarially trained using PGD attack with the same hyperparameters and the robust accuracy was measured based on AutoAttack. We note that for more complex datasets (CIFAR100 and Tiny ImageNet), this does not improve the robust accuracy for the uncompressed models. However, adversarial fine-tuning of compressed models does yield some improvement, particularly for the pruned models as reported in the $\mathcal{T}_{ad}(f^p_{st})$ column.

\begin{table}[t]
    \scriptsize
    \centering
    \caption{{Performance of compressed standard and robust models is evaluated on the MNIST, FashionMNIST, SVHN, CIFAR10, CIFAR100, and TinyImageNet datasets. We apply post-training quantization at \texttt{INT8} precision levels and $\ell_1$-pruning with 80\% sparsity ratio to WideResNet-50 and ViT architectures. After compressing the standardly trained models, we perform either standard fine-tuning, denoted as $\mathcal{T}_{st}(\cdot)$, or adversarial fine-tuning, denoted as $\mathcal{T}_{ad}(\cdot)$. Accuracy values are reported as ``$*/*$'', with the left value corresponding to standard accuracy and the right to robust accuracy by AutoAttack.}}
    \label{tab:final_quant}
    \makebox[1 \textwidth][c]{
        \begin{tabular}{cccccccc}
            \toprule
            \textbf{Dataset} & \textbf{Model} & $f_{st}$ & $f_{rb}$ & $\mathcal{T}_{st}(f_{st}^q)$ & $\mathcal{T}_{ad}(f_{st}^q)$ & $\mathcal{T}_{st}(f_{st}^p)$ & $\mathcal{T}_{ad}(f_{st}^p)$\\
            \midrule
            \multirow{2}{*}{MNIST} & {WRN} & 99.26/51.03 & \textbf{99.37}/92.24 & 
            99.07/53.82 & 99.10/\textbf{92.43} & 98.65/39.77& 98.88/91.82\\
            & {ViT} & \textbf{92.54}/31.56 & 92.01/\textbf{77.06} & 91.18/37.94 & 90.45/74.25 & 91.33/21.96& 91.01/74.08\\
            \midrule
            \multirow{2}{*}{FMNIST} & {WRN} & \textbf{91.20}/4.96 & 85.81/9.44&   89.72/6.60& 85.69/\textbf{11.25} & 89.05/5.66& 83.07/\textbf{9.56} \\
            & {ViT} & \textbf{85.22}/13.84 & 80.91/\textbf{24.22} &  84.62/16.53 & 79.17/22.10 & 82.17/14.45& 79.11/\textbf{24.44}\\
            \midrule
            \multirow{2}{*}{SVHN} & {WRN} & 90.42/6.51 & 89.51/38.37& 89.33/2.55 & \textbf{90.87}/\textbf{38.76} & 88.20/9.47& 88.29/35.46\\
            & {ViT} & 84.43/0.39 & 79.59/21.38 & \textbf{84.97}/0.29 & 81.50/\textbf{27.45} & 82.14/1.02 & 77.22/17.42\\
            \midrule
            \multirow{2}{*}{CIFAR10} & {WRN} & \textbf{86.48}/0.79& 75.92/19.99 & 85.84/3.91 & 80.88/\textbf{20.77} & 74.86/6.99& 63.58/26.88\\
            & {ViT} & \textbf{79.94}/1.15& 74.25/13.77 & 79.92/1.46 & 74.19/\textbf{16.71} & 69.45/8.73 & 61.69/21.05\\ 
            \midrule 
            \multirow{2}{*}{CIFAR100} & {WRN} & \textbf{64.87}/0.14& 55.11/3.09 & 62.28/1.32& 49.46/\textbf{4.09} & 47.17/6.31 & 38.57/\textbf{11.88}\\
            & {ViT} & \textbf{58.51}/0.78& 54.83/3.83& 51.71/0.43& 56.03/\textbf{4.92} & 42.62/5.91 & 36.57/\textbf{11.22}\\ 
            \midrule 
            \multirow{2}{*}{Tiny ImageNet} & {WRN} & \textbf{69.50}/0.15& 63.94/0.36 & 67.51/0.14 & 62.66/\textbf{0.41} & 65.35/5.69 & 62.98/\textbf{9.75}\\ %
            & {ViT} & \textbf{71.97}/0.02 & 67.66/0.35& 67.31/0.13& 68.84/\textbf{0.36} & 64.86/6.95& 63.61/\textbf{10.14}\\ 
            \bottomrule
        \end{tabular}
    }
    \vskip -0.15in
\end{table}

Generally, we observe that adversarial fine-tuning enables compressed models to recover their test performance, $f_{rb}^c$, and only slightly reduces it for compressed standard models, $f_{st}^c$. Both $f_{st}^c$ and $f_{rb}^c$ show notable gains in adversarial robustness after $\mathcal{T}_{ad}(\cdot)$. Importantly, adversarial fine-tuning of compressed standard models, $\mathcal{T}_{ad}(f_{st}^c)$, for only three epochs achieves robustness within about 5\% of the fully adversarially trained model $f_{rb}^c$. For example, pruned standard models after adversarial fine-tuning, $\mathcal{T}_{ad}(f_{st}^p)$, achieve robustness of $74.08$, whereas after standard fine-tuning, $\mathcal{T}_{st}(f_{st}^p)$, robustness was close to zero at $21.96$ for MNIST. Similarly, for quantized standard models, just three epochs of adversarial fine-tuning, $\mathcal{T}_{ad}(f_{st}^q)$, improved robustness from $37.94$ to $74.25$. A similar trend, though with smaller margins of improvement, is observed for Fashion-MNIST, SVHN, and CIFAR-10. These findings are consistent with~\cite{jeddi2020simple}, suggesting that a significant portion of the benefits of adversarial training can be obtained through minimal fine-tuning, even after compression.

Additional results for a wider range of quantization bit widths and pruning ratios are provided in \Cref{app:final}.


\section{Discussions}
\textbf{Adversarial fine-tuning instead of adversarial training:} Based on the experiments in~\Cref{sec:results}, we have shown that \emph{adversarial fine-tuning}, $\mathcal{T}_{ad}(\cdot)$, can substantially improve the robustness of \emph{compressed models}. With only three epochs of adversarial fine-tuning, robustness improves dramatically, from about $0\%$ to nearly the same levels as fully robust models. These gains are consistent across all datasets and both compression methods considered in this work, as shown in~\Cref{tab:final_quant}. We view this as a significant result, since the efficiency gains from compression and adversarial fine-tuning can be compounded. These experiments demonstrate that both efficiency and robustness can be jointly improved by applying adversarial fine-tuning to compressed models.

\noindent{\bf Pruning versus quantization:} Previous works comparing the test performance of pruning and quantization have often used compression ratios that may not be fair. For example, some compare models with a $50\%$ pruning ratio against models with \texttt{INT8} quantization precision~\cite{li2017pruning}. In our work, we performed a systematic tuning of compression levels for structured pruning and quantization to match their test performance, as shown in~Figure~\ref{fig:compressionlevels}. This resulted in the use of an $80\%$ sparsity ratio versus \texttt{INT8} precision for the Fashion-MNIST dataset. Furthermore, consistent with prior literature, we find that pruning depends on fine-tuning to recover test performance, whereas quantization does not necessarily benefit from fine-tuning.

\noindent{\bf Adversarial robustness of factorized neural networks:} We further extend 
\begin{table}[h]
    \vskip -0.2in
    \scriptsize
    \centering
    \caption{Performance of compressed standard and robust models on Fashion-MNIST dataset. We consider tensor decomposition with 50\% compression ratio for 8-layer CNN. After compression, we consider further performing standard fine-tuning $\mathcal{T}_{st} (\cdot)$, adversarial fine-tuning $\mathcal{T}_{ad} (\cdot)$, and without fine-tuning.}
    \label{tab:tensor}
    \begin{tabular}{ccccc}
        \toprule
        \multirow{3}{*}{\textbf{Model}} &\multirow{3}{*}{\textbf{Performance}} & \multicolumn{3}{c}{\textbf{Fine-Tuning}}\\
        \cmidrule{3-5}
        && \multicolumn{1}{c}{None} & \multicolumn{1}{c}{$\mathcal{T}_{st}(\cdot)$}& \multicolumn{1}{c}{$\mathcal{T}_{ad}(\cdot)$}\\
        \midrule
        \multirow{3}{*}{$f_{st}^d$} & test & 24.73 & \textbf{86.07} & 81.20 \\
        \cmidrule{2-5}
        & robustness & 3.05 & 3.60 & \textbf{75.40} \\
        \midrule
        \multirow{3}{*}{$f_{rb}^d$} & test & 30.75 & \textbf{85.30} & 81.83 \\
        \cmidrule{2-5}
        & robustness & 24.11 & 12.02 & \textbf{76.21} \\
        \bottomrule
    \end{tabular}
    \vskip -0.15in
\end{table}
our analysis to another common class of neural network compression methods. Let $f_{st}^d$ and $f_{rb}^d$ denote the decomposed standard and robust models, respectively, and adopt the same experimental settings as in pruning and quantization. \Cref{tab:tensor} reports the results on the Fashion-MNIST dataset, demonstrating that our observations also hold for factorized compression methods, including those based on singular value decomposition and tensor decomposition~\cite{novikov2015tensorizing,wangexploring}.

\noindent{\bf Computational gains:}
In our experiments we have shown that robustness can be achieved by fine-tuning of {\em compressed} models with only three epochs. Performing adversarial fine-tuning instead of adversarial training can reduce the computation time from about $118$ minutes to only about $14$ minutes on the CIFAR10 dataset. Furthermore, adversarial fine-tuning of {\em compressed} models is cheaper than fine-tuning of baseline models, and yields further reduction in computation time. For CIFAR10, we estimated that adversarial fine-tuning of a compressed model required around $10$ minutes. This indicates that the gains in computational efficiency are compounded when adversarial fine-tuning is performed on compressed models while retaining reasonable test and robustness performance, as shown in~\Cref{tab:fintuning}.

\noindent{\bf Limitations:}
We have performed multiple experiments to highlight the key results about the influence of adversarial fine-tuning of compressed neural networks. However, there still remain some limitations to our work and future extensions. 

In all our experiments we found fine-tuning for three epochs was adequate to improve the robustness performance. The number of fine-tuning epochs might be task-, dataset-, and model- dependent and should be carefully treated as another hyperparameter. This resulted in sub-optimal adversarial training of CIFAR100 and Tiny ImageNet. Furthermore, we did not perform any cross-architectural experiments on the two datasets. For instance, training ResNet-18 on Fashion-MNIST could allow us to explore to what extent a relatively more complex network can maintain robustness after compression. Conversely, the trade-off between efficiency and test performance when using a smaller network on CIFAR10 could also shine some light on the influence of using models with less scope for pruning. 



\section{Conclusion}
In this work, we set out to explore the interplay between model compression, test performance, and adversarial robustness. We have shown that adversarial fine-tuning of compressed models can yield robustness performance that is comparable to models that are adversarially trained from scratch.

With adversarial fine-tuning, the robustness performance of standard models is close to that of robust models. Our results across different neural networks and datasets suggest that adversarial fine-tuning might be a lighter substitute for adversarial training even when used alongside compression techniques like neural network pruning, quantization, or factorization. For PTQ with adversarial fine-tuning, all results have less than a 5\% point distance for both test and robustness performance between the standard and robust models.


In general, robust models perform better on both standard and adversarial performance measures. Adversarial fine-tuning does lend itself as an approach with lightweight training, for cases where less energy consumption and speed is favored over a marginal increase in performance. This yields a joint improvement of robustness and compute efficiency, as fine-tuning for a handful of epochs is considerably cheaper than full adversarial training. Based on these results, we conclude that we can obtain compressed models that are both efficient {\em and} robust.

\begin{credits}

\subsubsection{\ackname} The authors acknowledge funding received under European Union’s Horizon Europe Research and Innovation programme under grant agreements No. 101070284, No. 101070408 and No. 101189771.

\subsubsection{Disclosure of Interests.} The authors have no competing interests to declare that are relevant to the content of this article. 

\end{credits}
%
%
%
\bibliographystyle{splncs04}
\bibliography{main.bib}
\appendix
\newpage

\section{Experimental Set-up}
\subsection{Parameters for optimization during fine-tuning}\label{app:hyperparm}
After compression, the optimization hyperparameters are adjusted for both standard and adversarial fine-tuning. For pruning, the learning rate is increased to 0.1. For both PTQ and QAT, a momentum of 0.9 is added, and the learning rate is fixed at 0.01.

\subsection{Implementation details for t-SNE visualization of features}
We use t-SNE embedding implemented in scikit-learns to perform the visualizations. We set the perplexity to 30 and learning rate to ``auto''. Before applying the embedding, the features of the three last layers of every model pair are flattened.

We also visualize the inputs, both on clean images and on the images attacked by PGD with respect to each model, which is why we end up with three different labels (and not four) for the input plots in the first column of Figure~\ref{fig:bag_features}.

\section{Additional Results}

\begin{table}[h]
    \vskip -0.3in
    \scriptsize
    \centering
    \caption{Baseline performance of standard and robust models over Fashion-MNIST and CIFAR10 datasets comparing their test performance and robustness. For Fashion-MNIST, we additionally consider standard model with adversarial fine-tuning $\mathcal{T}_{ad} (\cdot)$.}
    \label{tab:baseline}
    \begin{tabular}{cccc}
        \toprule
        \textbf{Dataset} & {\bf Model} & \textbf{Test} & \textbf{Robustness} \\
        \midrule
        \multirow{4}{*}{Fashion-MNIST} &$f_{st}$ & 90.49$\pm$0.22 & 4.26$\pm$2.36  \\  
        \cmidrule{2-4}
        & $f_{rb}$& 87.87$\pm$0.33 & 82.51$\pm$0.16 \\
        \cmidrule{2-4}
        & $\mathcal{T}_{ad}(f_{st})$ & 85.37$\pm$0.44 & 77.53$\pm$1.17 \\ 
        \midrule
        \multirow{3}{*}{CIFAR10} & $f_{st}$ & 88.74$\pm$0.00 & 0.05$\pm$0.00 \\
        \cmidrule{2-4}
        & $f_{rb}$ & 85.72$\pm$0.27 & 57.22 $\pm$0.91\\
        \bottomrule
    \end{tabular}
    \vskip -0.5in
\end{table}

\subsection{Performance of compressed models on Fashion-MNIST without fine-tuning}
\begin{figure}[h]
    \centering
    \begin{subfigure}{0.48\textwidth}
        \centering
        \includegraphics[width=\linewidth]{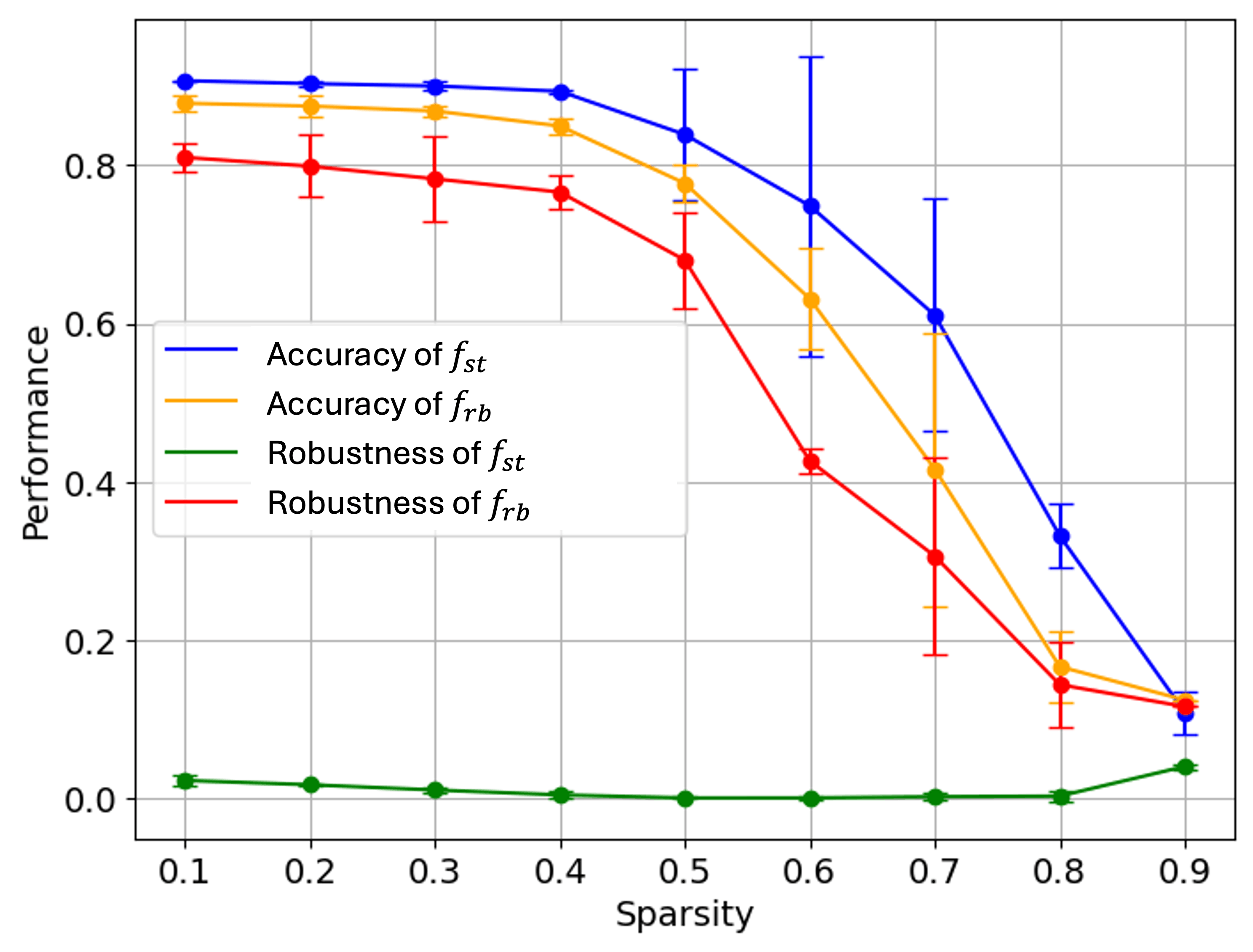}
        \caption{Standard/robust model with pruning}
        \label{fig:suba_nofinetune}
    \end{subfigure}
    \hfill
    \begin{subfigure}{0.48\textwidth}
        \centering
        \includegraphics[width=\linewidth]{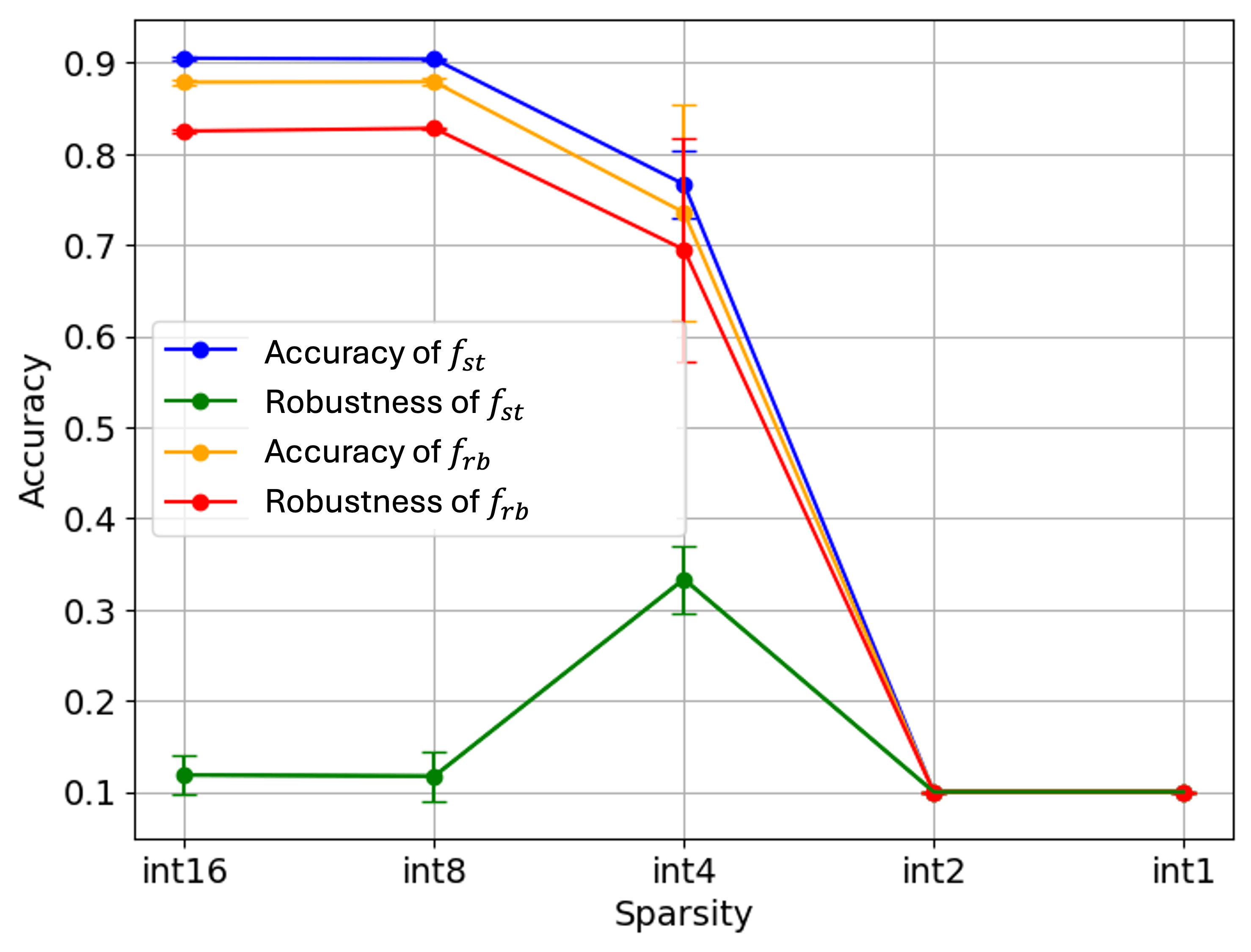}
        \caption{Standard/robust model with PTQ}
        \label{fig:subb_nofinetune}
    \end{subfigure}
    \caption{Performance of 8-layer compressed CNN models on Fashion-MNIST without fine-tuning. We perform $\ell_1$-norm pruning ($f^p$, left) and post-train quantization ($f^q$, right) on standard and robust models. In each subfigure, the horizontal axis shows the level of compression performed on the model, and the vertical axis shows the performance. Each model was trained three times and averages out, error bars show the standard deviation between runs.}
    \label{fig:nofinetune}
\end{figure}

We evaluate the test and robust performance of standard and robust models with different compression levels. Adversarial training is still an essential and effective way of improving the robustness performance of compressed models, as shown in \Cref{fig:nofinetune}.

\subsection{Performance of quantized robust models using QAT}
\begin{figure}[h]
    \centering
    \begin{subfigure}{0.48\textwidth}
        \centering
        \includegraphics[width=\linewidth]{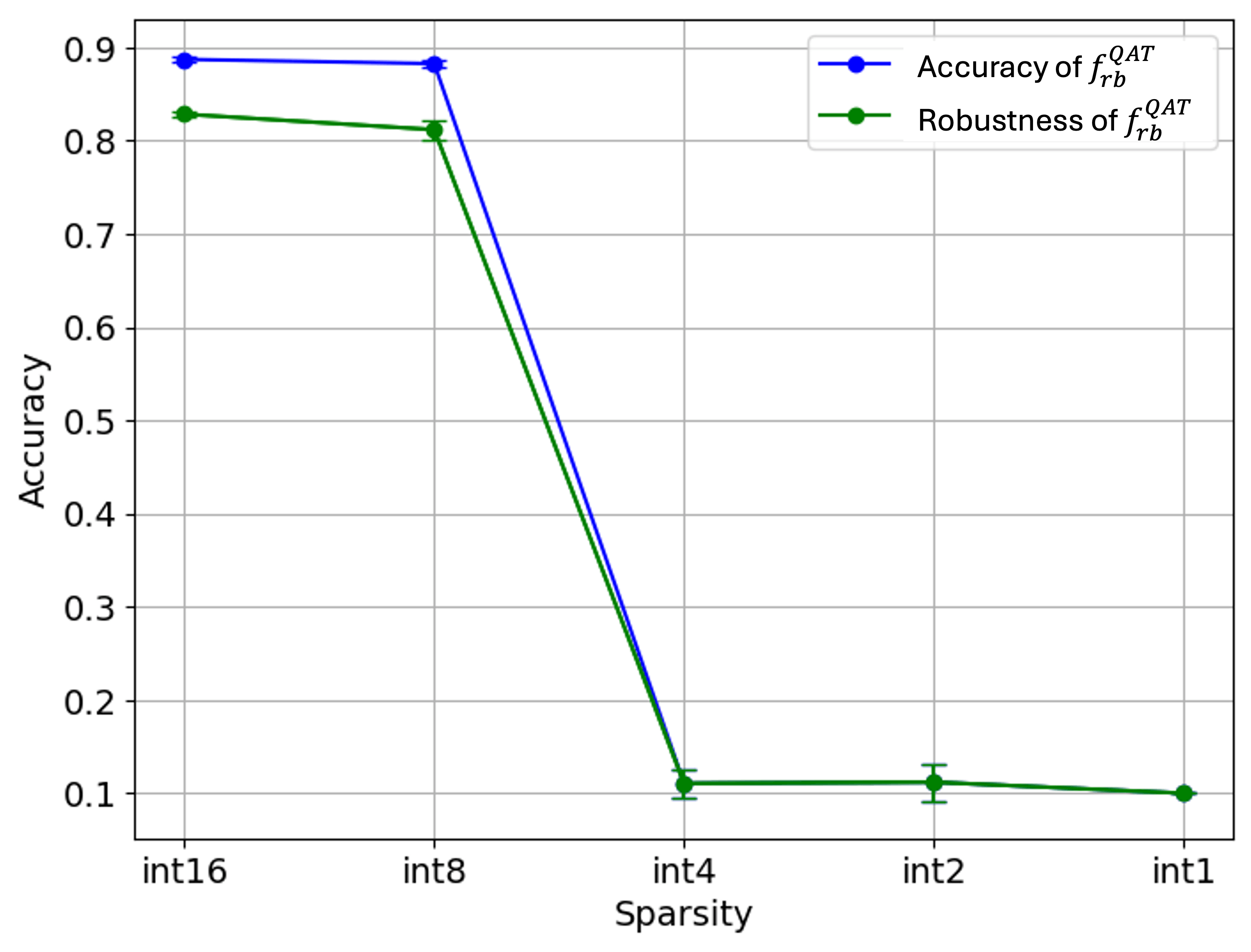}
        \caption{Robust model on F-MNIST with QAT}
        \label{fig:sub1a}
    \end{subfigure}
    \hfill
    \begin{subfigure}{0.48\textwidth}
        \centering
        \includegraphics[width=\linewidth]{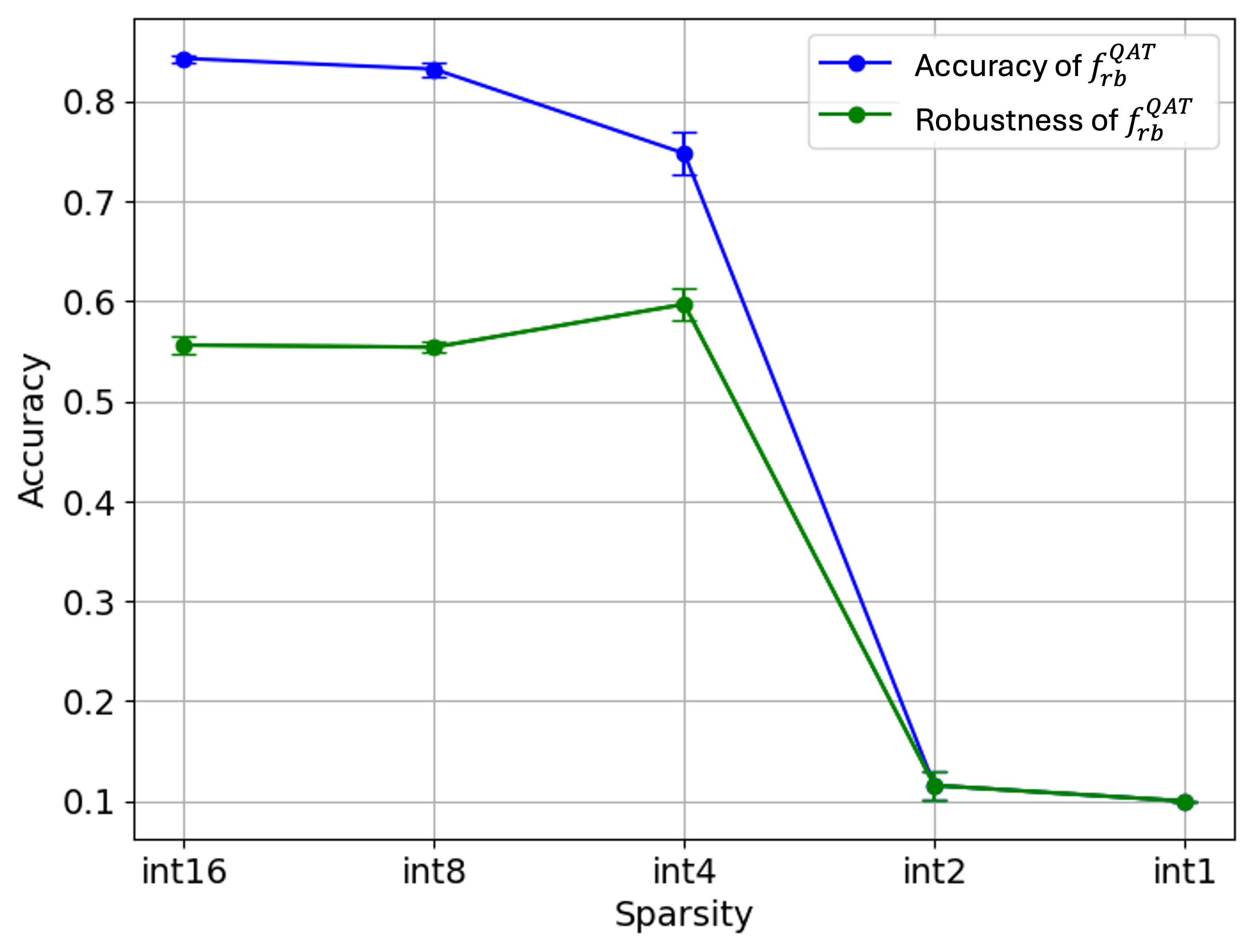}
        \caption{Robust model on CIFAR10 with QAT}
        \label{}
        \end{subfigure}
        \caption{Performance of 8-layer compressed 8-layer CNN on Fasion-MNIST ($f^q_{rb}$, left) and ResNet-18 on CIFAR10 ($f^q_{rb}$, right) without fine-tuning. We perform quantization-aware training with different precision on robust models. In each subfigure, the horizontal axis shows the level of compression performed on the model, and the vertical axis shows the performance. Each model was trained three times and averages out, error bars show the standard deviation between runs.}
        \label{fig:QATfine}    
\end{figure}

We test the robustness performance of a quantized robust model $f^q_{rb}$ with QAT. For Fashion-MNIST, we adversarially train the model from scratch with QAT, whereas for CIFAR10 we adversarially train on top of the pre-trained model ResNet-18. Our experiment reveal that in line with our comparison framework, the test performance among the various compression schemes remains highly similar, with differences of less than 5\% points, as shown in \Cref{fig:QATfine}.

\subsection{With versus without adversarial fine-tuning}
This section provides parallel experimental results of \Cref{tab:fintuning}, where we evaluate the effectiveness of adversarial fine-tuning on compressed models.

\begin{table}[h]
    \vskip -0.3in
    \scriptsize
    \centering
    \caption{Performance of 8-layer CNN on Fashion-MNIST dataset. For standard models, we consider the model $f_{st}$ without compression, the pruned model $f^p_{st}$ with 80\% sparsity ratio, the quantized model $f^q_{st}$ with \texttt{INT8} post-train quantization. For robust models, we consider the model $f_{rb}$ without compression, the pruned model $f^p_{rb}$ with 80\% sparsity ratio, the quantized model $f^q_{rb}$ with \texttt{INT8} post-train quantization and quantization-aware training. All compressed models are adversarially fine-tuned $\mathcal{T}_{ad} (\cdot)$.}
    \label{tab:fashionresults}
    \begin{tabular}{lcc}
        \toprule
        \textbf{Model} & \textbf{Test} & \textbf{Robustness}\\
        \midrule
        $f_{st}$ & 90.49$\pm$0.22 & 4.26$\pm$2.36 \\
        $\mathcal{T}_{ad} (f^p_{st})$ & 83.91 $\pm$1.53 & 76.74 $\pm$2.23 \\
        $\mathcal{T}_{ad} (f^q_{st})$ (PTQ) & \textbf{84.93$\pm$0.71}  & \textbf{79.43$\pm$0.59}  \\
        \midrule
        $f_{rb}$ & 87.87$\pm$0.33 & 82.51$\pm$0.16 \\
        $\mathcal{T}_{ad} (f^p_{rb})$ & 84.53$\pm$1.21 & 78.84$\pm$2.04 \\
        $\mathcal{T}_{ad} (f^q_{rb})$ (PTQ) & 87.51$\pm$0.04 & \textbf{82.65$\pm$0.11} \\ 
        $\mathcal{T}_{ad} (f^q_{rb})$ (QAT) & \textbf{88.27$\pm$0.42} & 81.18$\pm$1.08 \\
        \bottomrule
    \end{tabular}
    \vskip -0.3in
\end{table}
\noindent\textbf{Fashion-MNIST:} Examining the results on the Fashion-MNIST dataset as depicted in Table \ref{tab:fashionresults}, we find that with adversarial fine-tuning, the standard model demonstrates comparable performance to the robust model in terms of both test and robustness performance. This similarity can be attributed to the relatively straightforward nature of the Fashion-MNIST dataset, where robustness property are less intricate compared to more complex datasets. Notably, PTQ emerges as the highest performing method, achieving an robustness performance of 82.65\%. Even though QAT takes much longer to train, it does not seem to perform better than PTQ in our specific setting, and has a higher standard error. However, QAT does slightly outperform PTQ on test performance.

\begin{table}[h]
    \vskip -0.3in
    \scriptsize
    \centering
    \caption{Performance of ResNet-18 on CIFAR10 dataset. For standard models, we consider the model $f_{st}$ without compression, the pruned model $f^p_{st}$ with 50\% sparsity ratio, the quantized model $f^q_{st}$ with \texttt{INT8} post-train quantization. For robust models, we consider the model $f_{rb}$ without compression, the pruned model $f^p_{rb}$ with 50\% sparsity ratio, the quantized model $f^q_{rb}$ with \texttt{INT8} post-train quantization and quantization-aware training. All compressed models are adversarially fine-tuned $\mathcal{T}_{ad} (\cdot)$.} 
    \label{tab:cifarresults}
    \begin{tabular}{lcc}
        \toprule
        \textbf{Model} & \textbf{Test} & \textbf{Robustness}\\
        \midrule
        $f_{st}$ & 88.74$\pm$0.00 & 0.00$\pm$0.00 \\
        $\mathcal{T}_{ad} (f^p_{st})$ & 82.62$\pm$0.17 & 56.56$\pm$0.77 \\
        $\mathcal{T}_{ad} (f^q_{st})$ (PTQ) & \textbf{84.21$\pm$0.93} & \textbf{60.03$\pm$0.67}  \\
        \midrule
        $f_{rb}$ & 85.77$\pm$0.95 & 57.93$\pm$0.27 \\
        $\mathcal{T}_{ad} (f^p_{rb})$ & 83.55$\pm$0.86 & \textbf{57.27$\pm$0.47} \\
        $\mathcal{T}_{ad} (f^q_{rb})$ (PTQ) & \textbf{84.31$\pm$0.22} & 57.23$\pm$0.63 \\ 
        $\mathcal{T}_{ad} (f^q_{rb})$ (QAT) & 83.19$\pm$0.69 & 55.38$\pm$0.52 \\
        \bottomrule
    \end{tabular}
    \vskip -0.3in
\end{table}

\noindent\textbf{CIFAR10:} Analyzing the results on the CIFAR10 dataset presented in Table \ref{tab:cifarresults}, we see similar results as the Fashion-MNIST. After adversarial fine-tuning of the baseline models the test performance is reclaimed with difference of less than 5\% points. Additionally the standard model performs as well as the robust model with only 3 epochs of adversarial fine-tuning. The model without compression, the pruned model and the quantized model all achieve robust performance of $57.50\pm0.75$, showing again the effectiveness of adversarial fine-tuning. Surprisingly, PTQ outperforms QAT in on both test and robust performance.

Even though the benefits of QAT are not revealed in the results of \Cref{tab:fashionresults} and \Cref{tab:cifarresults}, we see that when performing QAT on a much more over-parameterized network, eg., ResNet-18 on CIFAR10, it better retains both test and robust performances when being quantized to \texttt{INT4}, see \Cref{fig:QATfine}. However, for the 8-layer CNN on Fashion-MNIST, QAT with \texttt{INT4} precision does not seem to work at all, as shown in \Cref{fig:compressionlevels}.

\subsection{Robust and non-robust features after compression}
To better characterize the influence of fine-tuning on compressed models, we analyze the intermediate feature maps of CNN models. We hypothesize that visualizing these feature maps can provide insights into how test performance and robustness are recovered through adversarial fine-tuning.

\begin{figure}[h]
    \centering
    \def\tsnesize{0.24}
    \begin{subfigure}{\tsnesize\textwidth}
        \centering
        \includegraphics[width=\linewidth]{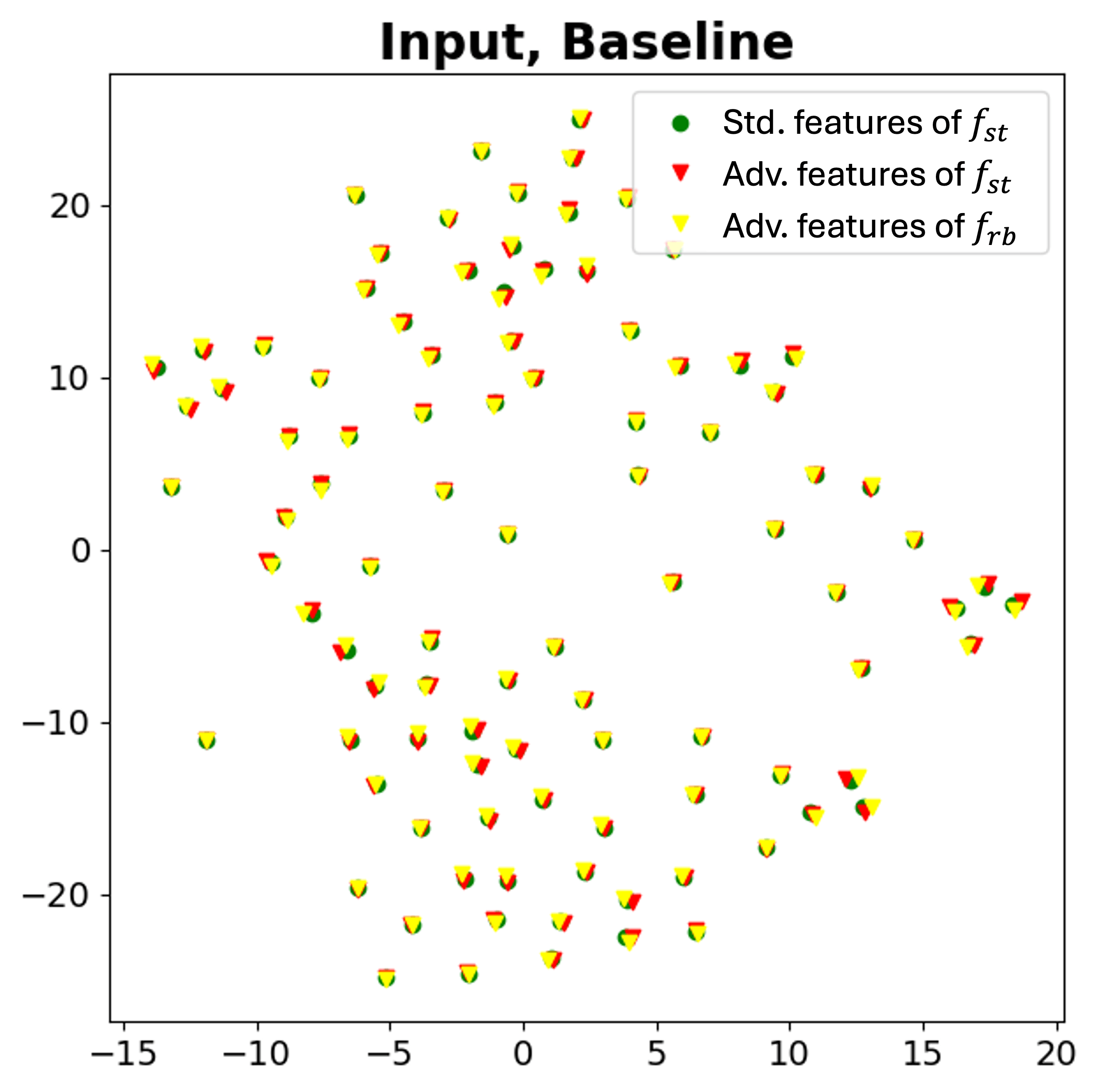}
    \end{subfigure}
    \hfill
    \begin{subfigure}{\tsnesize\textwidth}
        \centering
        \includegraphics[width=\linewidth]{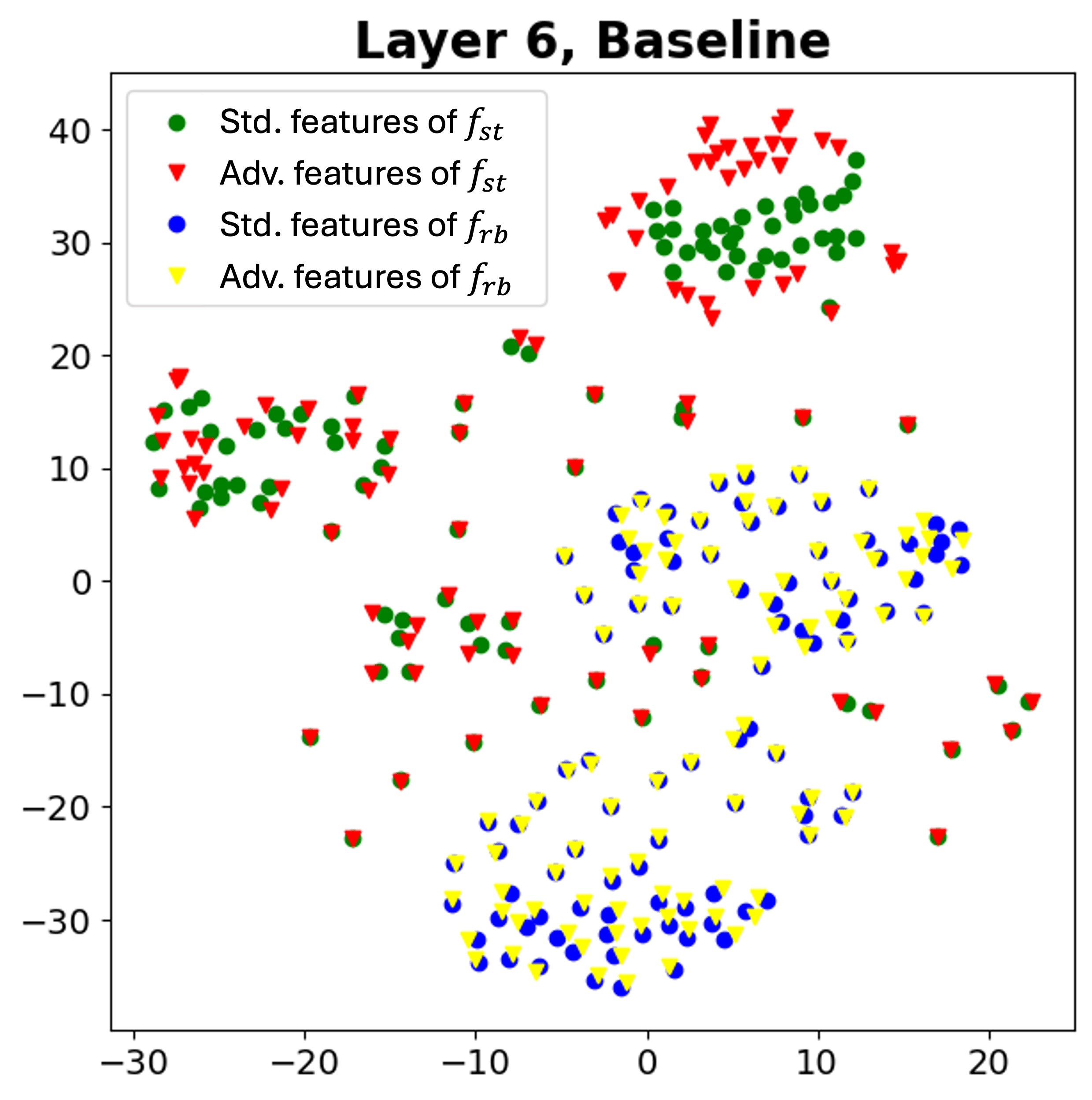}
    \end{subfigure}
    \hfill
    \begin{subfigure}{\tsnesize\textwidth}
        \centering
        \includegraphics[width=\linewidth]{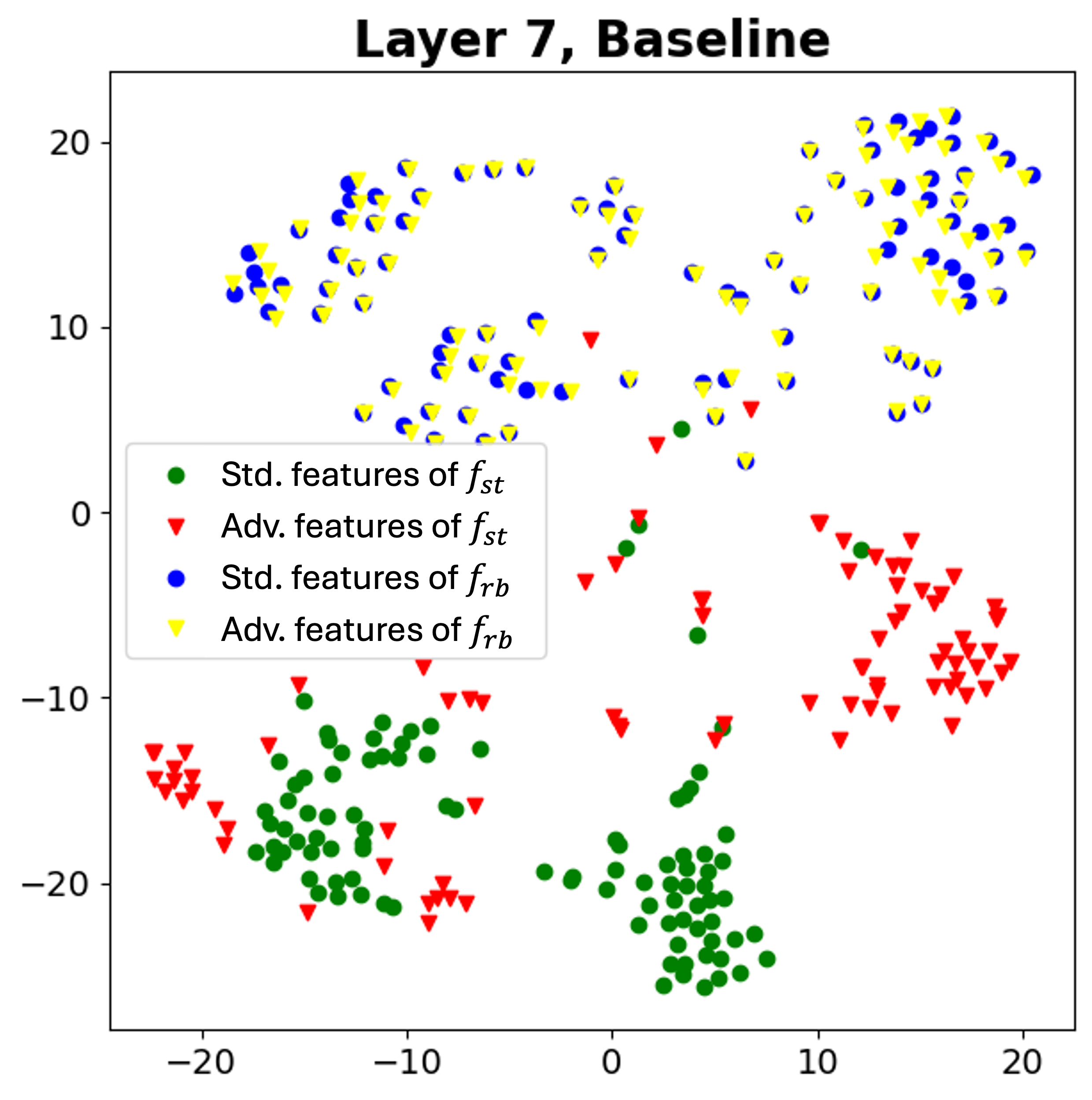}
    \end{subfigure}
    \hfill
    \begin{subfigure}{\tsnesize\textwidth}
        \centering
        \includegraphics[width=\linewidth]{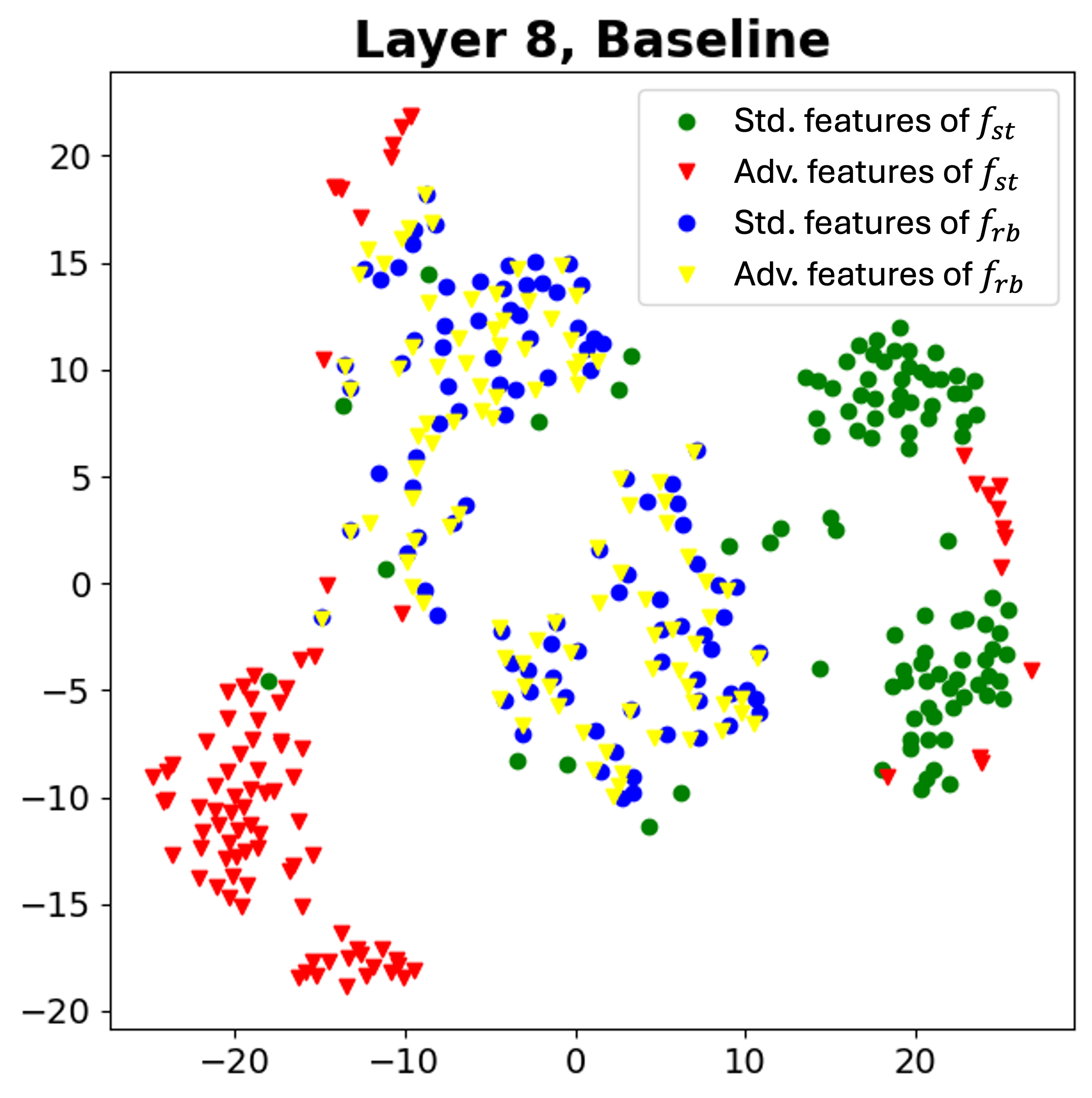}
    \end{subfigure}
    \begin{subfigure}{\tsnesize\textwidth}
        \centering
        \includegraphics[width=\linewidth]{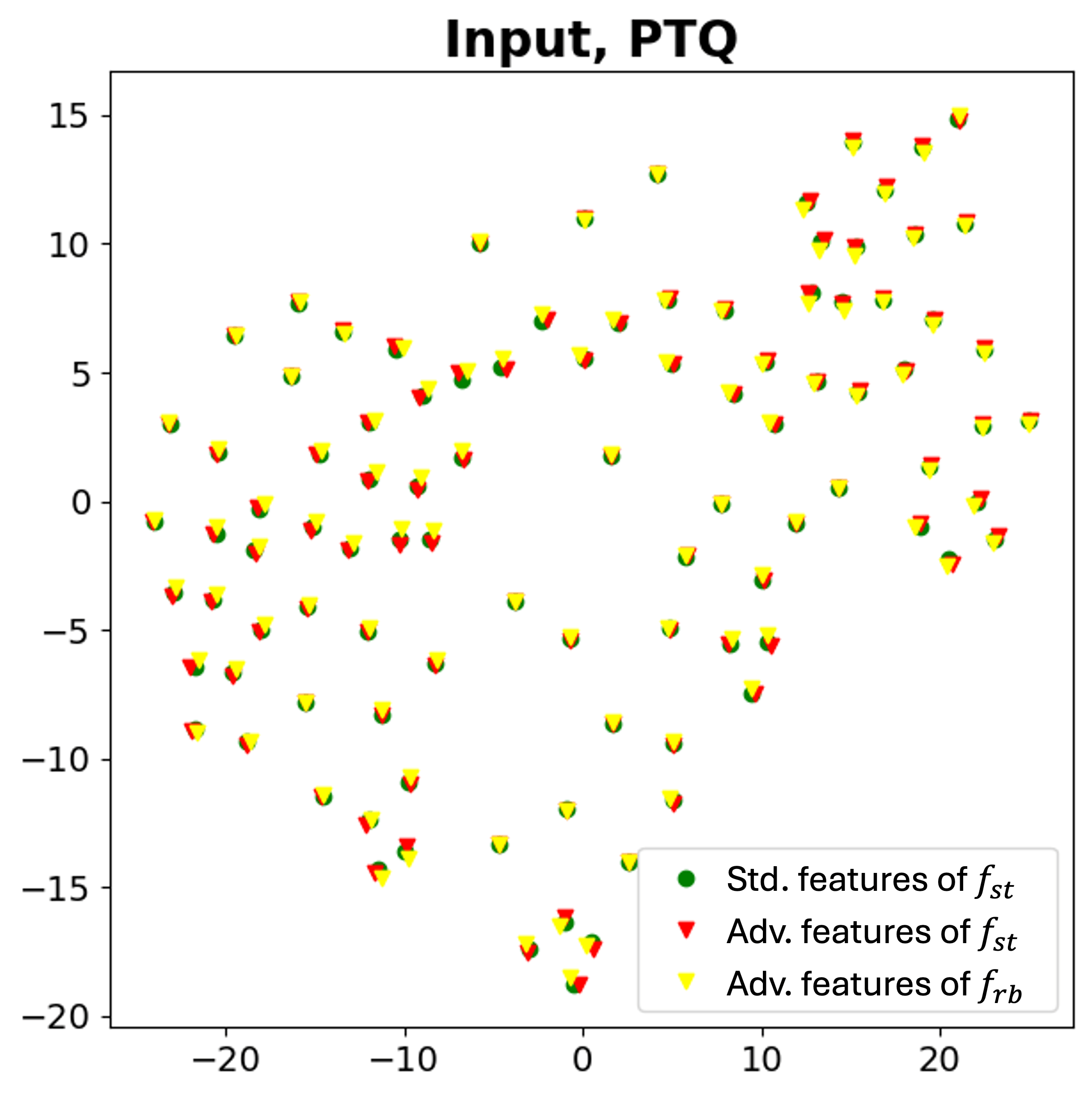}
    \end{subfigure}
    \hfill
    \begin{subfigure}{\tsnesize\textwidth}
        \centering
        \includegraphics[width=\linewidth]{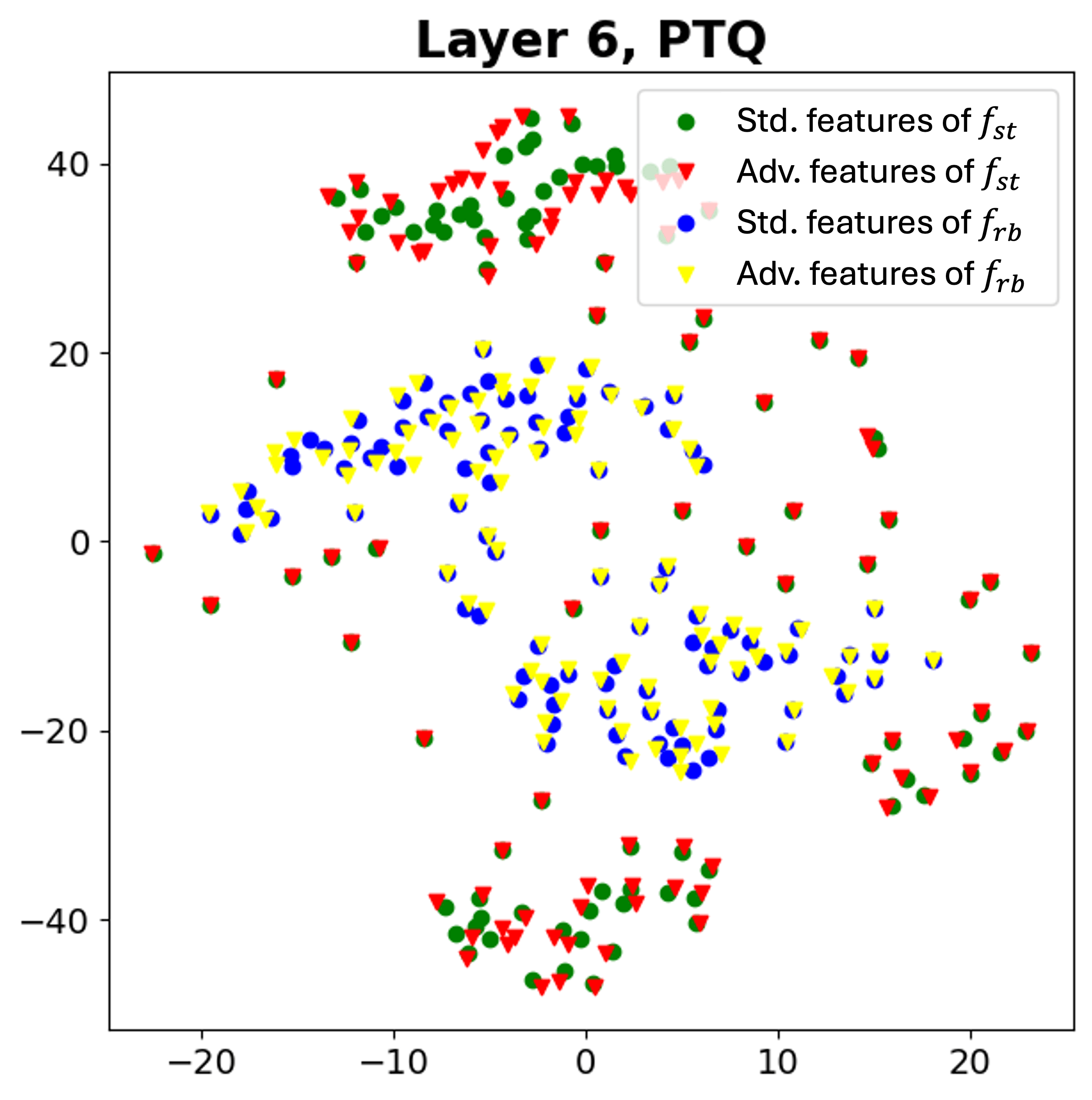}
    \end{subfigure}
    \hfill
    \begin{subfigure}{\tsnesize\textwidth}
        \centering
        \includegraphics[width=\linewidth]{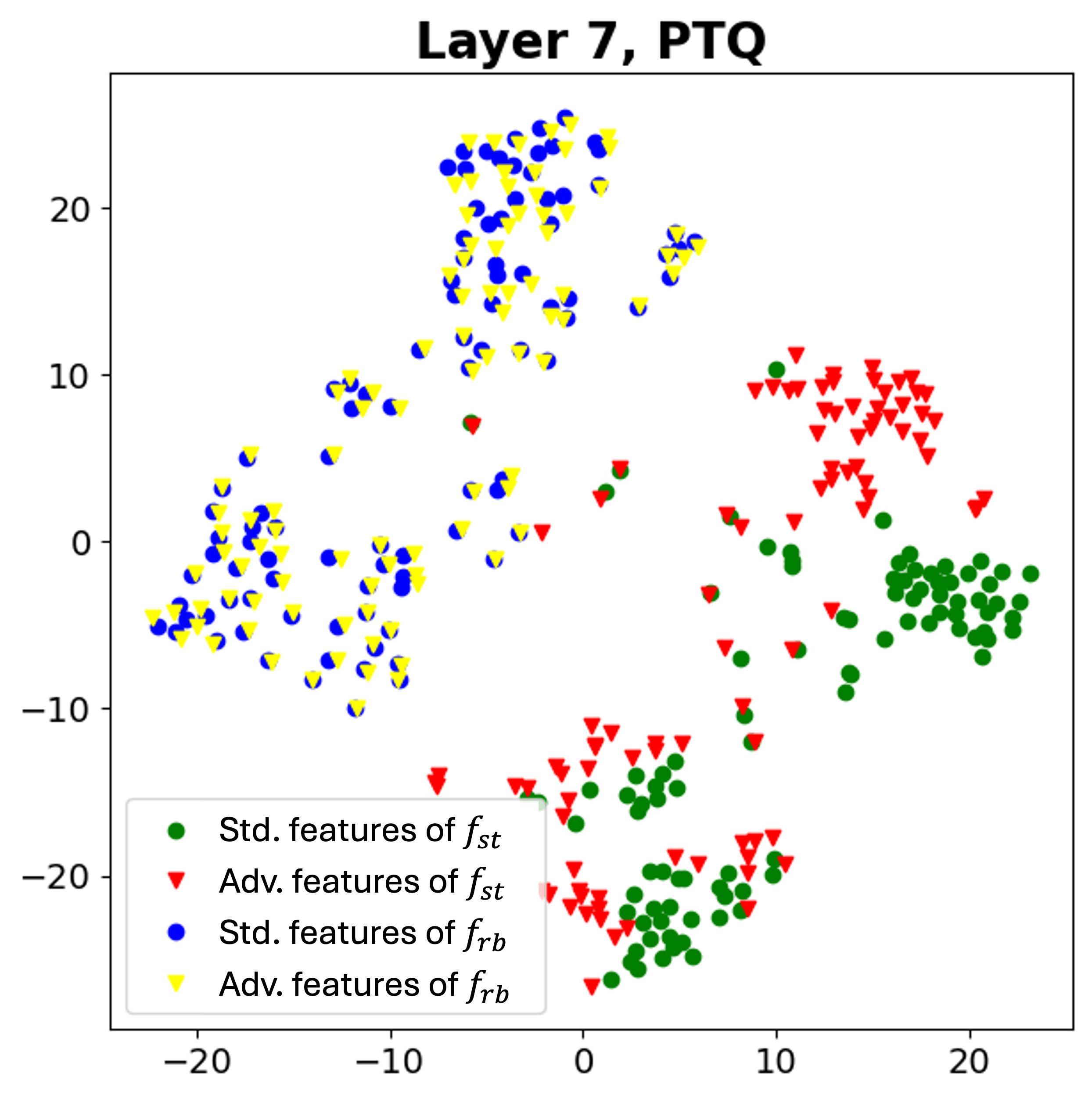}
    \end{subfigure}
    \hfill
    \begin{subfigure}{\tsnesize\textwidth}
        \centering
        \includegraphics[width=\linewidth]{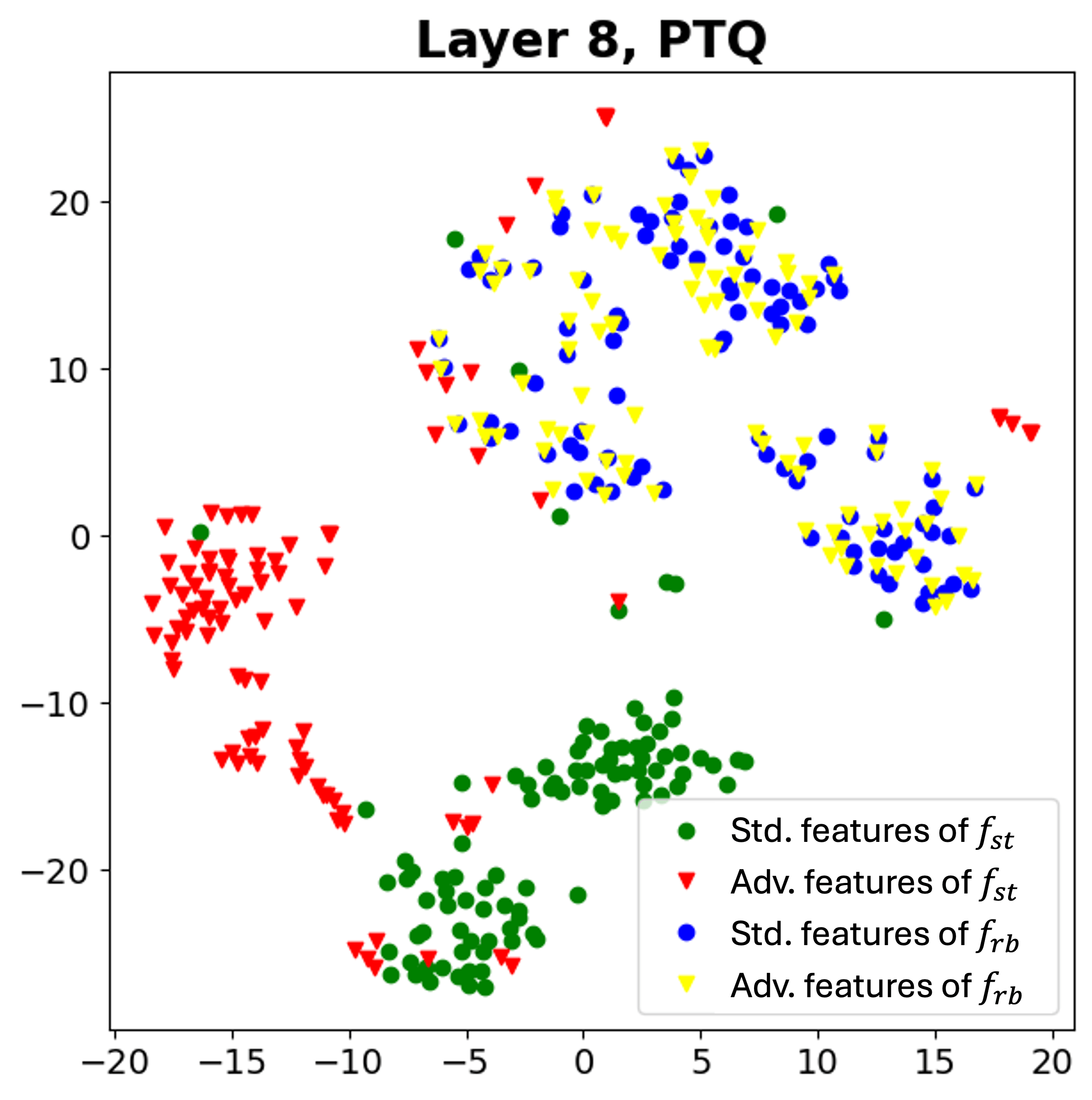}
    \end{subfigure}
    \begin{subfigure}{\tsnesize\textwidth}
        \centering
        \includegraphics[width=\linewidth]{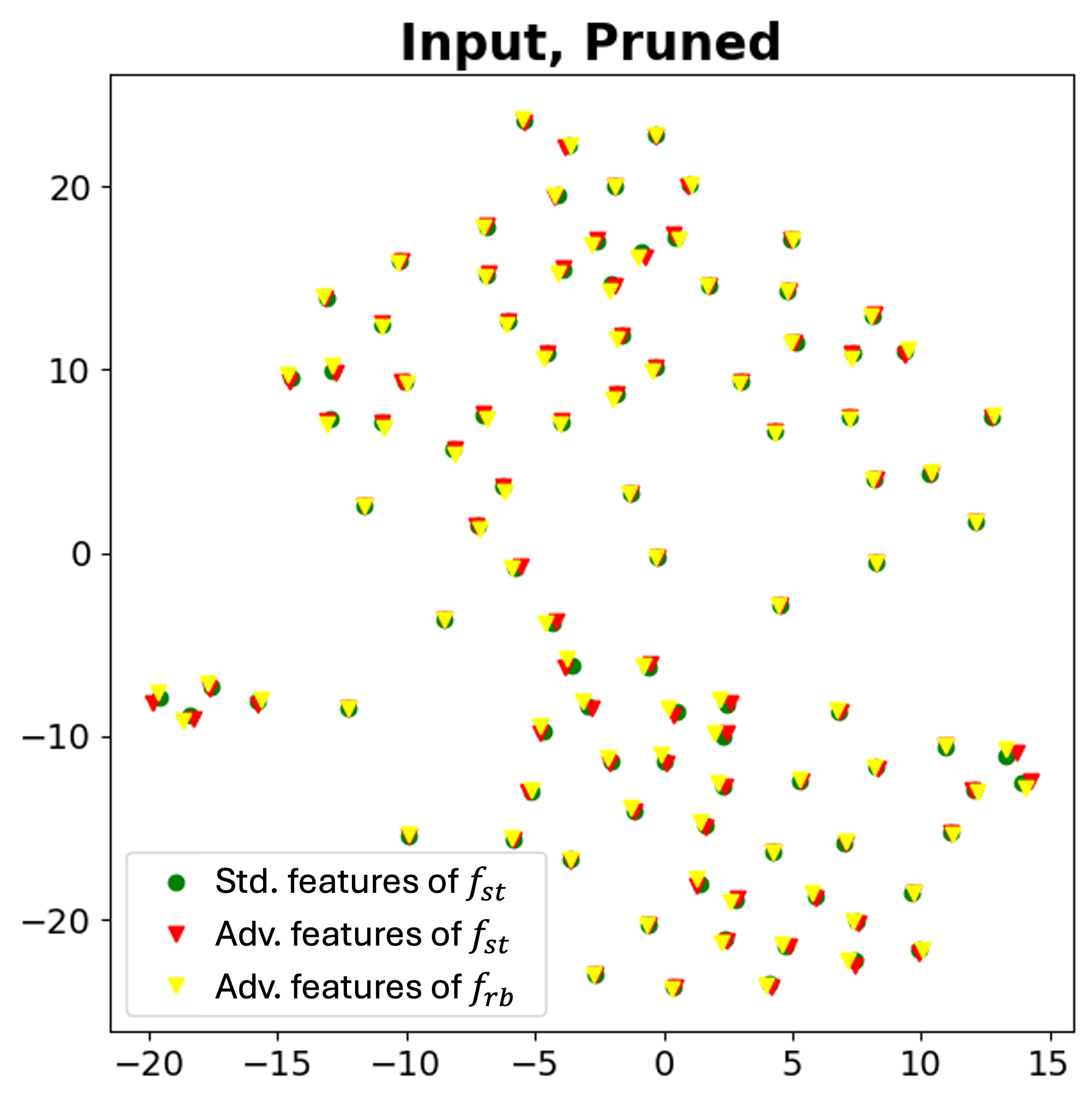}
    \end{subfigure}
    \hfill
    \begin{subfigure}{\tsnesize\textwidth}
        \centering
        \includegraphics[width=\linewidth]{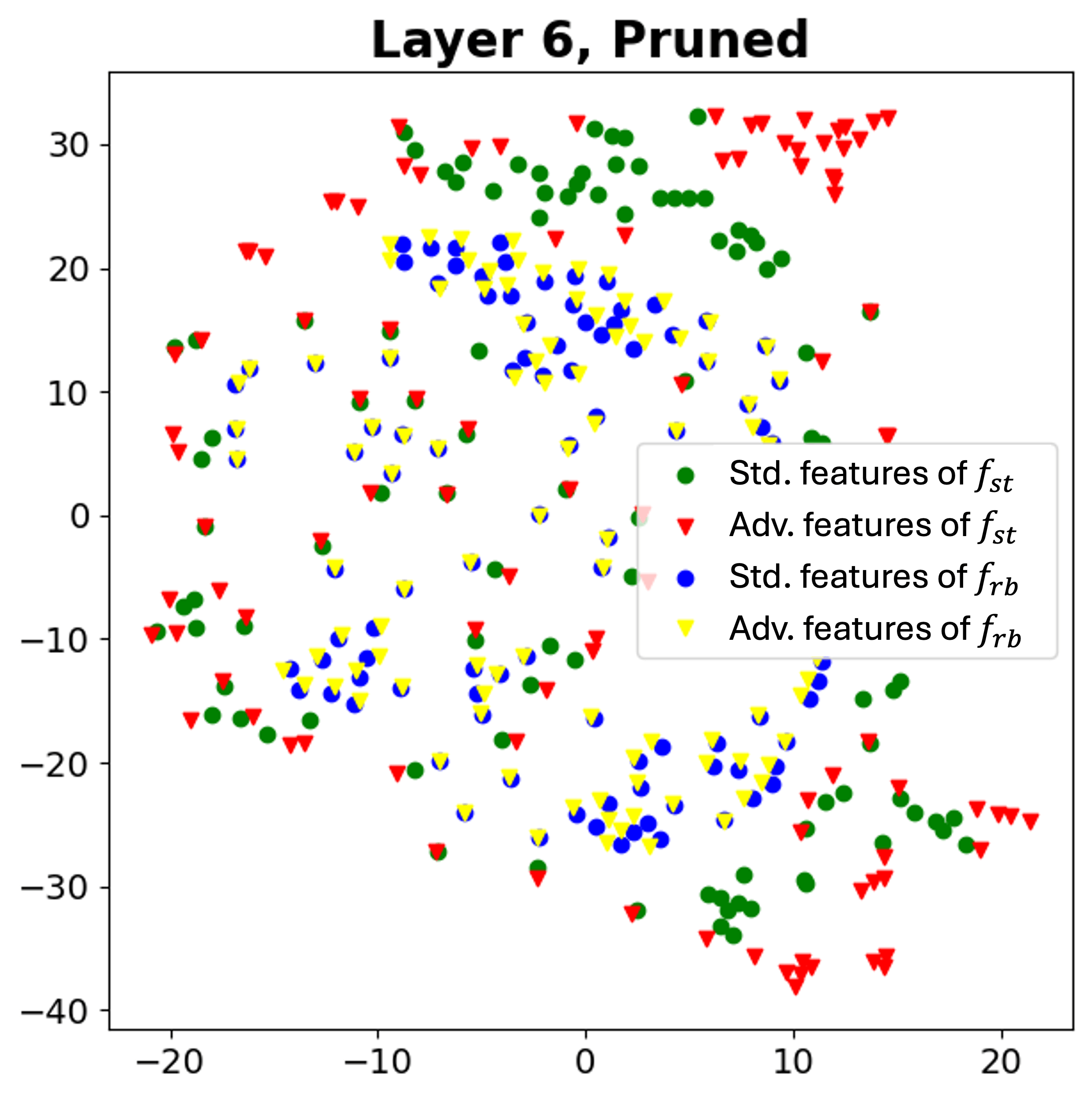}
    \end{subfigure}
    \hfill
    \begin{subfigure}{\tsnesize\textwidth}
        \centering
        \includegraphics[width=\linewidth]{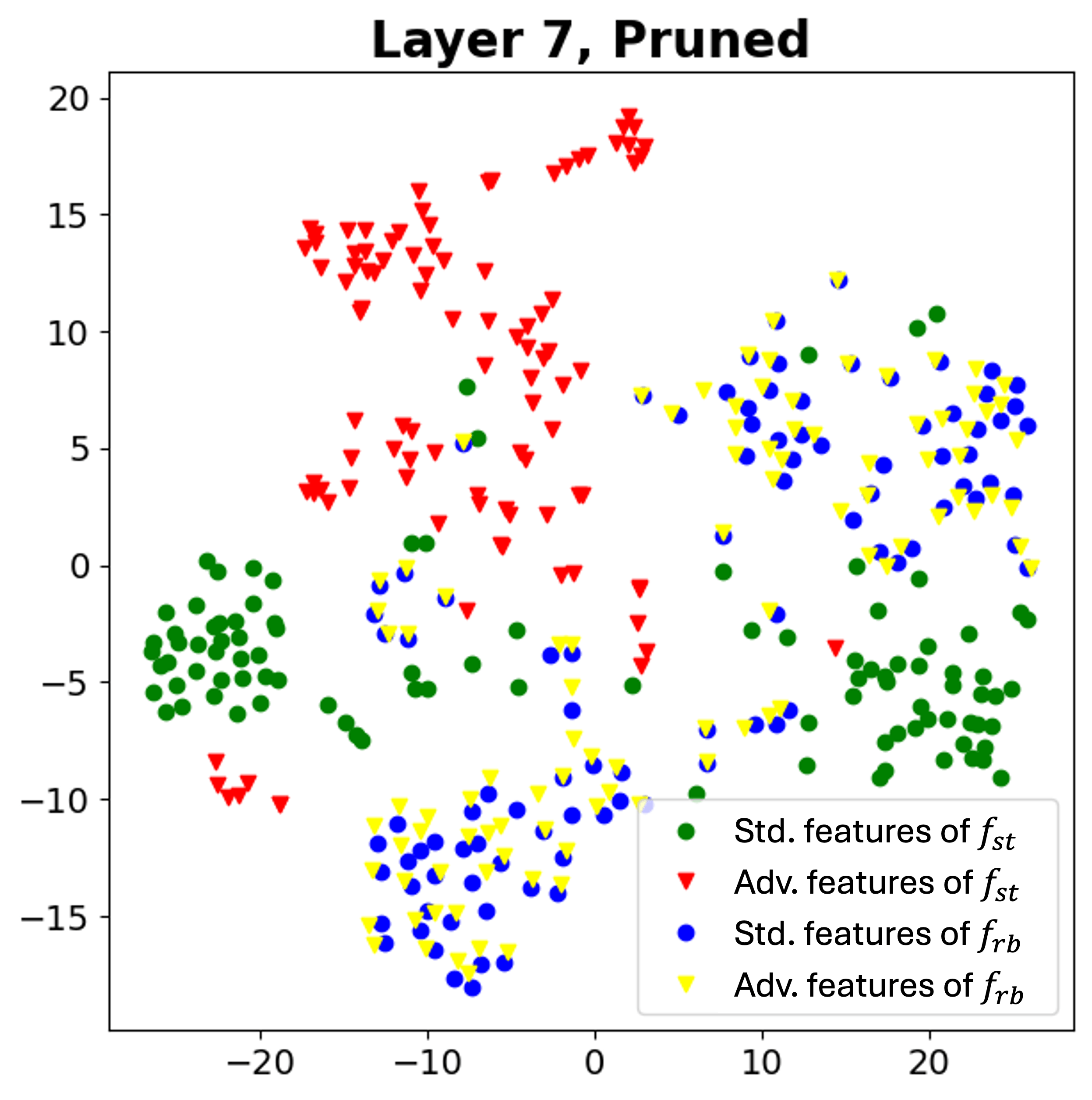}
    \end{subfigure}
    \hfill
    \begin{subfigure}{\tsnesize\textwidth}
        \centering
        \includegraphics[width=\linewidth]{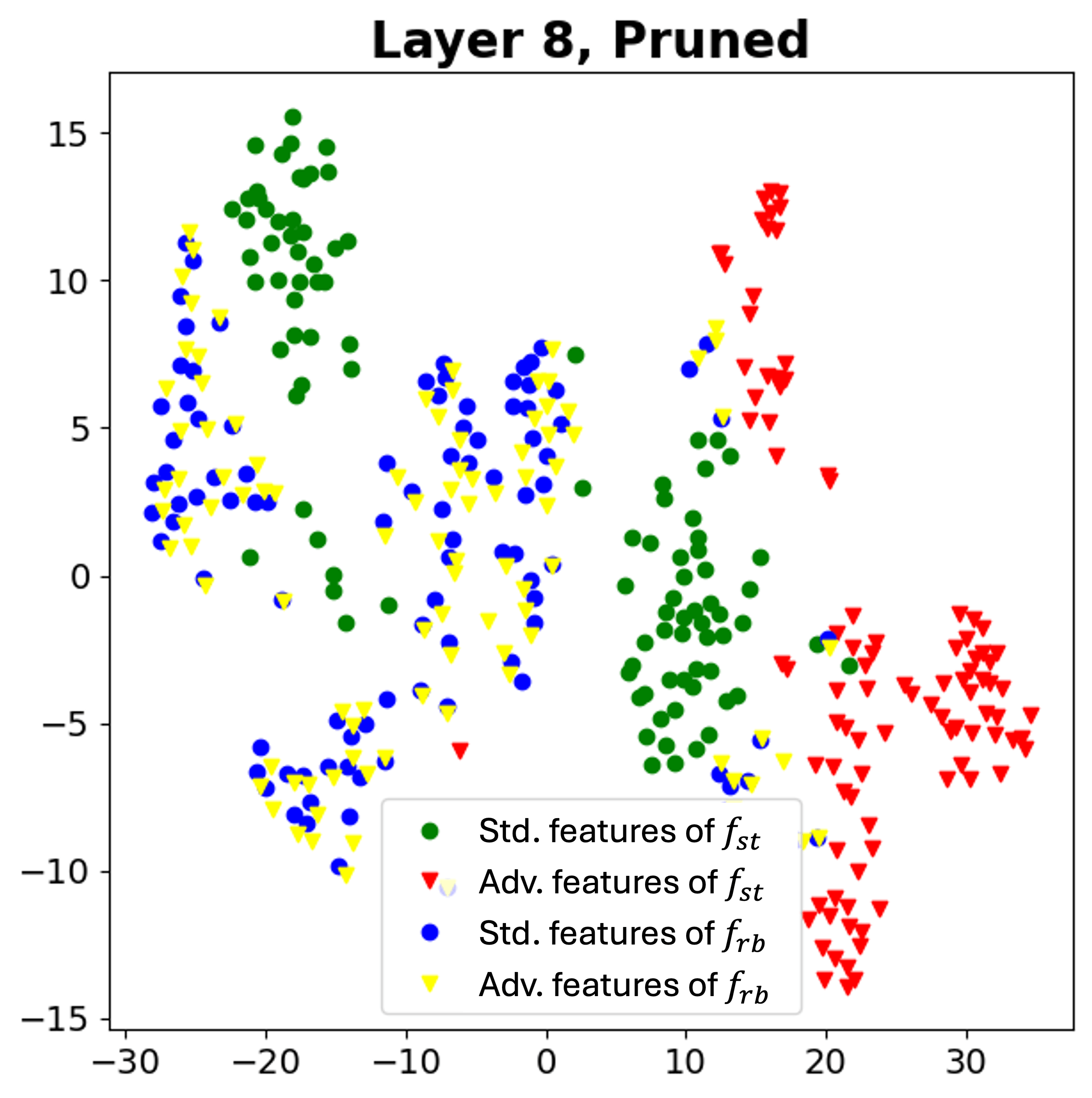}
    \end{subfigure}
        \caption{Features created by a 8-layer CNN on the subset of Fashion-MNIST dataset with class ``bag''. The first column shows t-SNE visualization generated from standard and adversarial images from white box attacks on the standard and robust models.. The last three columns show the features generated by the last three hidden layers (layer 6, 7, 8) of three different model pairs: standard and robust baseline models ($f_{st}$ versus $f_{rb}$), quantized (with \texttt{INT8} post-train quantization) standard models with and without adversarial fine-tuning ($f^q_{st}$ versus $\mathcal{T}_{ad} (f^q_{st})$), and pruned (with 80\% sparsity) standard model with standard and adversarial fine-tuning ($\mathcal{T}_{st} (f^p_{st})$ versus $\mathcal{T}_{ad} (f^p_{st})$).}
    \label{fig:bag_features}
\end{figure}

We use the intermediate feature maps of both standard and robust models. For clarity, we use our 8-layer CNN and evaluate it on Fashion-MNIST images from the ``bag'' class. The analysis is performed on three standard/robust model pairs: baseline, pruned, and quantized models. The t-SNE embeddings~\cite{van2008visualizing} of these feature maps are shown in~\Cref{fig:bag_features}.

The top row of~\Cref{fig:bag_features} shows the features produced by the standard and robust baseline models, $f_{st}$ and $f_{rb}$. The second row depicts the quantized (with PTQ) standard model after standard fine-tuning, $\mathcal{T}_{st}(f_{st}^q)$, and adversarial fine-tuning, $\mathcal{T}_{ad}(f_{st}^q)$. The bottom row shows the pruned (with 80\% sparsity ratio) standard model after standard fine-tuning, $\mathcal{T}_{st}(f_{st}^p)$, and adversarial fine-tuning, $\mathcal{T}_{ad}(f_{st}^p)$. Columns correspond to feature representations from the input layer and the 6th, 7th, and 8th hidden layers of the CNN.

In a typical CNN, when analyzing features for natural images, we often see distinct clusters representing different classes or patterns~\cite{zeiler2014visualizing}. However, when the model encounters adversarial examples, these clusters become less distinct and start to overlap. This effect is visible in~\Cref{fig:bag_features}, where the features of the standard models scatter in the later layers of the network. This suggests that misclassifications may not occur until the deeper layers, where more abstract features are considered. 

A striking property of robust features is their stability and consistency. They tend to remain in similar positions or maintain clear clustering in the feature space, regardless of adversarial perturbations. This consistency indicates that robust features are resilient to adversarial attacks. Moreover, our feature analysis highlights that robust models can classify both standard and adversarial images reliably. This observation extends to compressed models (the second and third rows in~\Cref{fig:bag_features}). The distinction between standard and adversarial images becomes clearer in the features extracted from the 6th, 7th, and 8th hidden layers of the models.

\subsection{Complete results of \Cref{tab:final_quant}} \label{app:final}
\begin{table}[h]
    \scriptsize
    \centering
    \caption{Performance of compressed standard and robust models is evaluated on the MNIST, FashionMNIST, SVHN, CIFAR10, CIFAR100, and TinyImageNet datasets. We apply post-training quantization at \texttt{INT16}, \texttt{INT8}, and \texttt{INT4} precision levels to WideResNet-50 and ViT architectures. After compressing the standardly trained models, we perform either standard fine-tuning, denoted as $\mathcal{T}_{st}(\cdot)$, or adversarial fine-tuning, denoted as $\mathcal{T}_{ad}(\cdot)$. Accuracy values are reported as ``$*/*$'', with the left value corresponding to standard accuracy and the right to robust accuracy.}
    \label{tab:final}
    \makebox[1 \textwidth][c]{
        \begin{tabular}{ccccccc}
            \toprule
            \textbf{Dataset} & \textbf{Model} & $f_{st}$ & $f_{rb}$ & \textbf{Bit}& $\mathcal{T}_{st}(f_{st}^q)$ & $\mathcal{T}_{ad}(f_{st}^q)$ \\
            \midrule
            \multirow{7}{*}{MNIST} & \multirow{3}{*}{WRN-50} & \multirow{3}{*}{99.26$\pm$0.04/51.03$\pm$1.22} & \multirow{3}{*}{99.37$\pm$0.04/92.24$\pm$0.07} & \texttt{INT16} & 99.14$\pm$0.04/51.36$\pm$0.54 & 99.28$\pm$0.04/92.67$\pm$0.14 \\
            &&&& \texttt{INT8} & 99.07$\pm$0.06/53.82$\pm$0.94 & 99.10$\pm$0.08/92.43$\pm$0.13 \\
            &&&& \texttt{INT4} & 95.22$\pm$0.92/58.06$\pm$1.24 & 92.84$\pm$2.40/90.39$\pm$2.36 \\
            \cmidrule{2-7}
            & \multirow{3}{*}{ViT} & \multirow{3}{*}{92.54$\pm$0.02/31.56$\pm$0.91} & \multirow{3}{*}{92.01$\pm$0.03/77.06$\pm$0.08} & \texttt{INT16} & 91.33$\pm$0.02/38.27$\pm$2.17 & 90.30$\pm$0.24/74.39$\pm$0.02 \\
            &&&& \texttt{INT8} & 91.18$\pm$0.53/37.94$\pm$2.33 & 90.45$\pm$0.15/74.25$\pm$0.07 \\
            &&&& \texttt{INT4} & 88.95$\pm$0.21/34.86$\pm$0.17 & 86.80$\pm$0.80/72.77$\pm$0.66 \\
            \midrule
            \multirow{7}{*}{FMNIST} & \multirow{3}{*}{WRN-50} & \multirow{3}{*}{91.20$\pm$0.26/4.96$\pm$0.93} & \multirow{3}{*}{85.81$\pm$0.21/9.44$\pm$0.14} & \texttt{INT16} & 90.38$\pm$0.14/6.71$\pm$1.82 & 85.37$\pm$0.32/11.34$\pm$0.74 \\
            &&&& \texttt{INT8} & 89.72$\pm$0.96/6.60$\pm$0.61 & 85.69$\pm$0.46/11.25$\pm$0.97 \\
            &&&& \texttt{INT4} & 88.07$\pm$0.19/6.27$\pm$0.15 & 85.71$\pm$5.76/10.68$\pm$7.09 \\
            \cmidrule{2-7}
            & \multirow{3}{*}{ViT} & \multirow{3}{*}{85.22$\pm$0.44/13.84$\pm$0.49} & \multirow{3}{*}{80.91$\pm$0.26/24.22$\pm$0.51} & \texttt{INT16} & 83.28$\pm$0.76/12.26$\pm$0.71 & 79.58$\pm$0.42/22.11$\pm$0.11 \\
            &&&& \texttt{INT8} & 84.62$\pm$0.06/16.53$\pm$0.99 & 79.17$\pm$0.32/22.10$\pm$0.30 \\
            &&&& \texttt{INT4} & 78.03$\pm$1.02/15.57$\pm$5.82 & 78.97$\pm$0.52/14.07$\pm$0.53 \\
            \midrule
            \multirow{7}{*}{SVHN} & \multirow{3}{*}{WRN-50} & \multirow{3}{*}{90.42$\pm$0.25/6.51$\pm$1.59} & \multirow{3}{*}{89.51$\pm$0.22/38.37$\pm$0.44} & \texttt{INT16} & 88.71$\pm$0.44/5.69$\pm$0.85 & 90.92$\pm$0.43/37.63$\pm$0.95 \\
            &&&& \texttt{INT8} & 89.33$\pm$0.41/2.55$\pm$0.81 & 90.87$\pm$0.29/38.76$\pm$0.29 \\
            &&&& \texttt{INT4} & 82.27$\pm$2.94/10.61$\pm$2.93 & 81.44$\pm$2.45/39.93$\pm$2.12 \\
            \cmidrule{2-7}
            & \multirow{3}{*}{ViT} & \multirow{3}{*}{84.43$\pm$2.17/0.39$\pm$0.21} & \multirow{3}{*}{79.59$\pm$0.51/21.38$\pm$1.30} & \texttt{INT16} & 85.31$\pm$0.39/0.28$\pm$0.30 & 83.88$\pm$2.29/24.16$\pm$4.57 \\
            &&&& \texttt{INT8} & 84.97$\pm$1.36/0.29$\pm$0.06 & 81.50$\pm$1.52/27.45$\pm$0.51 \\
            &&&& \texttt{INT4} & 84.70$\pm$0.36/3.56$\pm$0.22 & 82.80$\pm$3.32/20.10$\pm$0.80 \\
            \midrule
            \multirow{7}{*}{CIFAR10} & \multirow{3}{*}{WRN-50} & \multirow{3}{*}{68.63$\pm$0.39/0.63$\pm$0.11} & \multirow{3}{*}{60.24$\pm$1.48/15.89$\pm$0.51} & \texttt{INT16} & 63.74$\pm$2.78/2.27$\pm$0.17 & 65.70$\pm$1.95/15.21$\pm$2.44 \\
            &&&& \texttt{INT8} & 66.57$\pm$2.82/3.33$\pm$0.24 & 65.15$\pm$1.05/17.09$\pm$1.43 \\
            &&&& \texttt{INT4} & 62.39$\pm$3.75/3.83$\pm$0.34 & 61.17$\pm$2.57/10.74$\pm$0.42 \\
            \cmidrule{2-7}
            & \multirow{3}{*}{ViT} & \multirow{3}{*}{62.65$\pm$1.60/0.92$\pm$0.13} & \multirow{3}{*}{59.03$\pm$1.30/10.93$\pm$1.28} & \texttt{INT16} & 60.96$\pm$1.69/1.74$\pm$0.22 & 60.40$\pm$1.25/14.19$\pm$0.29 \\
            &&&& \texttt{INT8} & 63.47$\pm$1.15/1.16$\pm$0.23 & 58.94$\pm$2.26/13.25$\pm$0.79 \\
            &&&& \texttt{INT4} & 55.81$\pm$1.43/1.31$\pm$0.10 & 56.89$\pm$0.17/11.62$\pm$0.38 \\
            \midrule
            \multirow{7}{*}{CIFAR100} & \multirow{3}{*}{WRN-50} & \multirow{3}{*}{40.99$\pm$0.12/0.09$\pm$0.02} & \multirow{3}{*}{29.78$\pm$1.40/2.00$\pm$0.14} & \texttt{INT16} & 35.69$\pm$0.59/0.57$\pm$0.04 & 33.89$\pm$1.17/2.64$\pm$0.07 \\
            &&&& \texttt{INT8} & 34.65$\pm$1.26/0.73$\pm$0.08 & 33.78$\pm$0.89/2.78$\pm$0.11 \\
            &&&& \texttt{INT4} & 25.04$\pm$0.31/0.43$\pm$0.03 & 24.81$\pm$0.44/2.35$\pm$0.05 \\
            \cmidrule{2-7}
            & \multirow{3}{*}{ViT} & \multirow{3}{*}{36.95$\pm$3.47/0.50$\pm$0.04} & \multirow{3}{*}{34.61$\pm$0.23/2.48$\pm$0.33} & \texttt{INT16} & 33.88$\pm$3.47/0.50$\pm$0.04 & 34.33$\pm$3.07/3.24$\pm$0.19 \\
            &&&& \texttt{INT8} & 32.63$\pm$2.49/0.24$\pm$0.09 & 35.35$\pm$2.24/3.34$\pm$0.13 \\
            &&&& \texttt{INT4} & 26.50$\pm$1.28/0.74$\pm$0.06 & 31.53$\pm$0.56/2.13$\pm$0.04 \\
            \midrule
            \multirow{7}{*}{TImageNet} & \multirow{3}{*}{WRN-50} & \multirow{3}{*}{31.82$\pm$0.26/0.07$\pm$0.03} & \multirow{3}{*}{29.28$\pm$0.44/0.22$\pm$0.02} & \texttt{INT16} & 27.94$\pm$0.49/0.04$\pm$0.01 & 28.11$\pm$0.53/0.24$\pm$0.06 \\
            &&&& \texttt{INT8} & 26.47$\pm$1.06/0.07$\pm$0.02 & 26.80$\pm$0.08/0.25$\pm$0.03 \\
            &&&& \texttt{INT4} & 16.56$\pm$0.18/0.06$\pm$0.03 & 17.74$\pm$0.10/0.70$\pm$0.09 \\
            \cmidrule{2-7}
            & \multirow{3}{*}{ViT} & \multirow{3}{*}{32.93$\pm$0.18/0.01$\pm$0.00} & \multirow{3}{*}{30.99$\pm$0.49/0.23$\pm$0.05} & \texttt{INT16} & 27.72$\pm$0.21/0.06$\pm$0.01 & 30.84$\pm$0.68/0.22$\pm$0.07  \\
            &&&& \texttt{INT8} & 28.46$\pm$0.34/0.07$\pm$0.03 & 31.49$\pm$0.10/0.22$\pm$0.04 \\
            &&&& \texttt{INT4} & 26.07$\pm$0.01/0.11$\pm$0.02 & 25.03$\pm$0.37/0.14$\pm$0.02 \\
            \bottomrule
        \end{tabular}
    }
\end{table}

\begin{table}[h]
    \scriptsize
    \centering
    \caption{Performance of compressed standard and robust models is evaluated on the MNIST, FashionMNIST, SVHN, CIFAR10, CIFAR100, and TinyImageNet datasets. We apply $\ell_1$-pruning with 20\%, 50\%, 80\% sparsity ratio to WideResNet-50 and ViT architectures. After compressing the standardly trained models, we perform either standard fine-tuning, denoted as $\mathcal{T}_{st}(\cdot)$, or adversarial fine-tuning, denoted as $\mathcal{T}_{ad}(\cdot)$. Accuracy values are reported as ``$*/*$'', with the left value corresponding to standard accuracy and the right to robust accuracy.}
    \label{tab:final}
    \makebox[1 \textwidth][c]{
        \begin{tabular}{ccccccc}
            \toprule
            \textbf{Dataset} & \textbf{Model} & $f_{st}$ & $f_{rb}$ & \textbf{Ratio}& $\mathcal{T}_{st}(f_{st}^p)$ & $\mathcal{T}_{ad}(f_{st}^p)$ \\
            \midrule
            \multirow{7}{*}{MNIST} & \multirow{3}{*}{WRN-50} & \multirow{3}{*}{99.26$\pm$0.04/51.03$\pm$1.22} & \multirow{3}{*}{99.37$\pm$0.04/92.24$\pm$0.07} & \texttt{0.2} & 98.16$\pm$1.26/41.79$\pm$5.99 & 99.04$\pm$0.08/92.14$\pm$0.21 \\
            &&&& \texttt{0.5} & 98.58$\pm$0.57/47.09$\pm$4.52 & 98.96$\pm$0.20/92.26$\pm$0.31 \\
            &&&& \texttt{0.8} & 98.65$\pm$0.02/39.77$\pm$1.64 & 98.88$\pm$0.05/91.82$\pm$0.20 \\
            \cmidrule{2-7}
            & \multirow{3}{*}{ViT} & \multirow{3}{*}{92.54$\pm$0.02/31.56$\pm$0.91} & \multirow{3}{*}{92.01$\pm$0.03/77.06$\pm$0.08} & \texttt{0.2} & 90.57$\pm$1.26/23.03$\pm$5.99 & 91.43$\pm$0.10/76.81$\pm$0.30 \\
            &&&& \texttt{0.5} & 91.13$\pm$0.20/28.33$\pm$2.23 & 91.38$\pm$0.11/76.70$\pm$0.30 \\
            &&&& \texttt{0.8} & 91.33$\pm$0.02/21.96$\pm$1.64 & 91.01$\pm$0.10/74.08$\pm$0.35 \\
            \midrule
            \multirow{7}{*}{FMNIST} & \multirow{3}{*}{WRN-50} & \multirow{3}{*}{91.20$\pm$0.26/4.96$\pm$0.93} & \multirow{3}{*}{85.81$\pm$0.21/9.44$\pm$0.14} & \texttt{0.2} & 87.07$\pm$0.74/8.75$\pm$0.82 & 85.44$\pm$0.17/11.53$\pm$0.63 \\
            &&&& \texttt{0.5} & 87.04$\pm$0.83/7.46$\pm$3.07 & 85.02$\pm$0.41/11.47$\pm$0.08 \\
            &&&& \texttt{0.8} & 89.05$\pm$0.45/5.66$\pm$0.27 & 83.07$\pm$0.26/9.56$\pm$0.72 \\
            \cmidrule{2-7}
            & \multirow{3}{*}{ViT} & \multirow{3}{*}{85.22$\pm$0.44/13.84$\pm$0.49} & \multirow{3}{*}{80.91$\pm$0.26/24.22$\pm$0.51} & \texttt{0.2} & 80.43$\pm$0.97/19.25$\pm$0.96 & 79.17$\pm$0.68/27.88$\pm$0.70 \\
            &&&& \texttt{0.5} & 80.99$\pm$0.91/17.29$\pm$0.98 & 78.29$\pm$0.84/26.30$\pm$0.93 \\
            &&&& \texttt{0.8} & 82.17$\pm$0.60/14.45$\pm$0.80 & 79.11$\pm$0.85/24.44$\pm$0.76 \\
            \midrule
            \multirow{7}{*}{SVHN} & \multirow{3}{*}{WRN-50} & \multirow{3}{*}{90.42$\pm$0.25/6.51$\pm$1.59} & \multirow{3}{*}{89.51$\pm$0.22/38.37$\pm$0.44} & \texttt{0.2} & 88.80$\pm$1.06/12.35$\pm$1.58 & 89.43$\pm$0.49/36.77$\pm$0.26 \\
            &&&& \texttt{0.5} & 89.70$\pm$0.28/13.74$\pm$1.24 & 89.55$\pm$0.37/37.30$\pm$0.20 \\
            &&&& \texttt{0.8} & 88.20$\pm$0.27/9.47$\pm$0.83 & 88.29$\pm$0.21/35.46$\pm$0.66 \\
            \cmidrule{2-7}
            & \multirow{3}{*}{ViT} & \multirow{3}{*}{84.43$\pm$2.17/0.39$\pm$0.21} & \multirow{3}{*}{79.59$\pm$0.51/21.38$\pm$1.30} & \texttt{0.2} & 82.38$\pm$1.13/3.21$\pm$0.78 & 79.65$\pm$0.85/21.16$\pm$0.91 \\
            &&&& \texttt{0.5} & 83.71$\pm$0.46/2.62$\pm$0.72 & 79.63$\pm$2.77/20.31$\pm$1.13 \\
            &&&& \texttt{0.8} & 82.14$\pm$0.95/1.02$\pm$0.25 & 77.22$\pm$1.80/17.42$\pm$1.20 \\
            \midrule
            \multirow{7}{*}{CIFAR10} & \multirow{3}{*}{WRN-50} & \multirow{3}{*}{68.63$\pm$0.39/0.63$\pm$0.11} & \multirow{3}{*}{60.24$\pm$1.48/15.89$\pm$0.51} & \texttt{0.2} & 58.13$\pm$5.91/7.43$\pm$2.44 & 60.23$\pm$4.84/20.78$\pm$4.66 \\
            &&&& \texttt{0.5} & 61.38$\pm$7.65/8.85$\pm$4.65 & 57.76$\pm$5.01/20.35$\pm$4.62 \\
            &&&& \texttt{0.8} & 60.49$\pm$2.36/6.50$\pm$1.28 & 50.41$\pm$4.58/15.65$\pm$3.95 \\
            \cmidrule{2-7}
            & \multirow{3}{*}{ViT} & \multirow{3}{*}{62.65$\pm$1.60/0.92$\pm$0.13} & \multirow{3}{*}{59.03$\pm$1.30/10.93$\pm$1.28} & \texttt{0.2} & 57.25$\pm$3.17/8.60$\pm$1.45 & 56.91$\pm$3.30/22.60$\pm$2.40 \\
            &&&& \texttt{0.5} & 60.53$\pm$3.87/6.88$\pm$2.14 & 56.03$\pm$4.81/21.22$\pm$1.45 \\
            &&&& \texttt{0.8} & 56.92$\pm$2.10/5.35$\pm$1.25 & 51.80$\pm$2.50/18.40$\pm$1.85 \\
            \midrule
            \multirow{7}{*}{CIFAR100} & \multirow{3}{*}{WRN-50} & \multirow{3}{*}{40.99$\pm$0.12/0.09$\pm$0.02} & \multirow{3}{*}{29.78$\pm$1.40/2.00$\pm$0.14} & \texttt{0.2} & 34.19$\pm$0.54/2.11$\pm$0.14 & 31.22$\pm$1.47/7.82$\pm$1.28 \\
            &&&& \texttt{0.5} & 32.73$\pm$1.47/2.13$\pm$0.17 & 31.43$\pm$1.97/5.42$\pm$1.46 \\
            &&&& \texttt{0.8} & 29.82$\pm$1.95/1.25$\pm$0.16 & 23.16$\pm$1.58/4.98$\pm$1.22 \\
            \cmidrule{2-7}
            & \multirow{3}{*}{ViT} & \multirow{3}{*}{36.95$\pm$3.47/0.50$\pm$0.04} & \multirow{3}{*}{34.61$\pm$0.23/2.48$\pm$0.33} & \texttt{0.2} & 29.16$\pm$1.85/2.07$\pm$0.25 & 29.42$\pm$1.28/7.65$\pm$1.05 \\
            &&&& \texttt{0.5} & 29.56$\pm$2.53/1.84$\pm$0.29 & 28.55$\pm$1.04/6.73$\pm$1.52 \\
            &&&& \texttt{0.8} & 26.72$\pm$2.95/0.98$\pm$0.14 & 23.12$\pm$1.60/4.54$\pm$1.23 \\
            \midrule
            \multirow{7}{*}{TImageNet} & \multirow{3}{*}{WRN-50} & \multirow{3}{*}{31.82$\pm$0.26/0.07$\pm$0.03} & \multirow{3}{*}{29.28$\pm$0.44/0.22$\pm$0.02} & \texttt{0.2} & 25.46$\pm$0.72/0.33$\pm$0.03 & 29.86$\pm$0.79/9.38$\pm$0.55 \\
            &&&& \texttt{0.5} & 27.78$\pm$0.96/0.51$\pm$0.32 & 28.07$\pm$0.80/9.62$\pm$0.84 \\
            &&&& \texttt{0.8} & 26.86$\pm$0.71/2.42$\pm$0.19 & 23.92$\pm$0.92/7.47$\pm$0.90 \\
            \cmidrule{2-7}
            & \multirow{3}{*}{ViT} & \multirow{3}{*}{32.93$\pm$0.18/0.01$\pm$0.00} & \multirow{3}{*}{30.99$\pm$0.49/0.23$\pm$0.05} & \texttt{0.2} & 26.76$\pm$0.83/1.41$\pm$0.25 & 29.60$\pm$0.61/9.55$\pm$0.62  \\
            &&&& \texttt{0.5} & 28.88$\pm$0.96/2.62$\pm$0.33 & 29.17$\pm$0.80/9.74$\pm$0.84 \\
            &&&& \texttt{0.8} & 26.25$\pm$1.14/3.99$\pm$0.42 & 24.70$\pm$0.90/7.96$\pm$0.71 \\
            \bottomrule
        \end{tabular}
    }
\end{table}
\end{document}